\documentclass[12pt,twoside]{report}
 \pdfoutput=1 
\usepackage[utf8]{inputenc}  
\usepackage[numbers]{natbib}
\usepackage{amsmath}
\usepackage[linesnumbered,ruled,vlined]{algorithm2e}
\usepackage[noend]{algpseudocode}
\usepackage{float}
\usepackage{listings}
\usepackage{xcolor}\usepackage{setspace}
\usepackage{mathtools}
\usepackage{enumitem}
\usepackage{pifont}
\usepackage{mathabx}
\setcitestyle{square}
\usepackage{verbatim}
\usepackage{textcomp,    
            xspace}

\restylefloat{table}

\DeclareMathOperator*{\argmin}{arg\!\min}

\newcommand{\reportauthor}{Seyedeh Niusha Alavi Foumani}
\newcommand{\supervisor}{Professor Wayne Luk}
\newcommand{\degreetype}{Advanced Computing}

\usepackage[a4paper,hmargin=2.8cm,vmargin=2.0cm,includeheadfoot]{geometry}
\usepackage{textpos}
\usepackage{tabularx,longtable,multirow,subfigure,caption}
\usepackage{fancyhdr} 
\usepackage{url} 
\usepackage[english]{babel}
\usepackage{amsmath}
\usepackage{graphicx}
\usepackage{dsfont}
\usepackage{epstopdf} 
\usepackage{backref} 
\usepackage{array}
\usepackage{latexsym}
\usepackage[pdftex,pagebackref,hypertexnames=false,colorlinks]{hyperref} 

\hypersetup{pdftitle={},
  pdfsubject={}, 
  pdfauthor={},
  pdfkeywords={}, 
  pdfstartview=FitH,
  pdfpagemode={UseOutlines},
  bookmarksnumbered=true, bookmarksopen=true, colorlinks,
    citecolor=black,%
    filecolor=black,%
    linkcolor=black,%
    urlcolor=black}

\usepackage[all]{hypcap}

\def\@makechapterhead#1{%
  \vspace*{10\p@}%
  {\parindent \z@ \raggedright \sffamily
    \interlinepenalty\@M
    \Huge\bfseries \thechapter \space\space #1\par\nobreak
    \vskip 30\p@
  }}

\def\@makeschapterhead#1{%
  \vspace*{10\p@}%
  {\parindent \z@ \raggedright
    \sffamily
    \interlinepenalty\@M
    \Huge \bfseries  #1\par\nobreak
    \vskip 30\p@
  }}

\allowdisplaybreaks

\renewcommand{\vec}[1]{{\boldsymbol{{#1}}}} 
\newcommand{\mat}[1]{{\boldsymbol{{#1}}}} 

\date{September 2020}

\begin{document}


\begin{titlepage}
\newcommand{\HRule}{\rule{\linewidth}{0.5mm}} 


\includegraphics[width = 4cm]{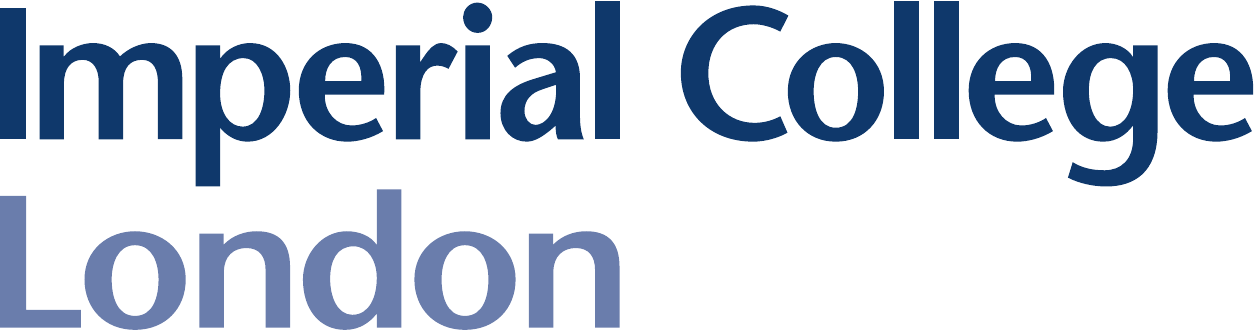}\\[0.5cm] 

\center 


\textsc{\Large Imperial College London}\\[0.5cm] 
\textsc{\large Department of Computing}\\[0.5cm] 


\HRule \\[0.4cm]
{ \Large \bfseries An FPGA Accelerated Method for Training Feed-forward Neural Networks Using Alternating Direction Method of Multipliers and LSMR \\[0.3cm]} 
\HRule \\[1.5cm]
 

\begin{minipage}{0.4\textwidth}
\begin{flushleft} \large
\emph{Author:}\\
\reportauthor 
\end{flushleft}
\end{minipage}
~
\begin{minipage}{0.4\textwidth}
\begin{flushright} \large
\emph{Supervisor:} \\
\supervisor 
\end{flushright}
\end{minipage}\\[4cm]

\vfill 
Submitted in partial fulfilment of the requirements for the MSc degree in
\degreetype~of Imperial College London\\[0.5cm]

\makeatletter
\@date 
\makeatother

\end{titlepage}

\pagenumbering{roman}
\clearpage{\pagestyle{empty}\cleardoublepage}
\setcounter{page}{1}
\pagestyle{fancy}
\renewcommand{\baselinestretch}{1.25}\normalsize
\begin{abstract}
In this project, we have successfully designed, implemented, deployed and tested a novel FPGA accelerated algorithm for neural network training. The algorithm itself was developed in an independent study option. This training method is based on Alternating Direction Method of Multipliers algorithm, which has strong parallel characteristics and avoids procedures such as matrix inversion that are problematic in hardware designs by employing LSMR. As an intermediate stage, we fully implemented the ADMM-LSMR method in C language for feed-forward neural networks with a flexible number of layers and hidden size. We demonstrated that the method can operate with fixed-point arithmetic without compromising the accuracy. Next, we devised an FPGA accelerated version of the algorithm using Intel FPGA SDK for OpenCL and performed extensive optimisation stages followed by successful deployment of the program on an Intel Arria 10 GX FPGA. The FPGA accelerated program showed up to 6 times speed up comparing to equivalent CPU implementation while achieving promising accuracy.\vspace{8pt}\\


\end{abstract}

\cleardoublepage
\section*{Acknowledgements}
I would like to thank the following people:
\begin{itemize}[label=$\sqbullet$]
    \item My supervisor, Prof Wayne Luk for all his valuable help and advice and for giving me the opportunity to get involved in an exciting area of research.
    \item Dr Ce Guo for all of his guidance and constructive feedback. He introduced the fascinating world of hardware for neural networks to me and patiently taught me a lot throughout the course of this project.
    \item My family and my fiancé for their unconditional love and support.
    
\end{itemize}


\clearpage{\pagestyle{empty}\cleardoublepage}

\fancyhead[RE,LO]{\sffamily {Table of Contents}}
\doublespacing
\tableofcontents
\renewcommand{\baselinestretch}{1.25}\normalsize

\clearpage{\pagestyle{empty}\cleardoublepage}
\pagenumbering{arabic}
\setcounter{page}{1}
\newcommand{\changefont}{%
    \fontsize{7}{7}\selectfont}
\fancyhead[LE,RO]{\changefont\slshape \rightmark \tiny}
\fancyhead[LO,RE]{\changefont\slshape \leftmark \small}

\chapter{Introduction}
\label{intro}

In this project, we have designed and implemented a hardware-accelerated neural network training algorithm. This project is, in fact, a continuation of an independent study option, in which a hardware-friendly approach for training neural networks using ADMM and LSMR was introduced \cite{iso}. In this work, we present an implementation of the ADMM-LSMR algorithm, which is accelerated with FPGA using OpenCL. To the best of our knowledge, this is the first hardware-accelerated implementation of an ADMM-based training method that uses LSMR to avoid matrix inversion. This implementation takes advantage of parallel characteristics of ADMM and LSMR and uses fixed-point arithmetic to suit hardware design restrictions.

\section{Motivation}
 Machine learning and in particular neural networks have shown promising performance in many domains both in academia and industry. The models are getting more and more complex, and the available amount of data is rapidly growing. As a result, more sophisticated challenges are emerging in this field, and many techniques have been employed to make training algorithms more efficient and keep up with the data volume and complexity. One way to address some of these challenges is to fortify the training platforms by employing hardware acceleration or even designing custom hardware.\vspace{8pt}\\
 Several approaches can be taken in order to use hardware acceleration in neural network training algorithms. Gradient-based methods, which are a commonly used category of algorithms for training neural networks, have many characteristics that complicate the use of hardware acceleration. In \cite{iso}, the ADMM-LSMR method is described as an alternative to gradient-based methods alongside a high-level Python implementation as proof of concept. This method is, in fact, orthogonal to common approaches as it is focused on the algorithm itself to be suitable for hardware design rather than implementing a hardware-accelerated variant of a conventional training algorithm. However, a method being theoretically suitable for hardware design and a Python implementation is far away from a practical and deployable hardware design.
 \section{Objectives}
 In this project, we improved the proposed method in \cite{iso} by applying some common techniques and made an even more hardware friendly variant of the algorithm. Then we took multiple steps to optimise the hardware design and finally, a fully operational deployment on an FPGA card using OpenCL was achieved. \vspace{8pt}\\
 The key components of this project and the achievements can be enumerated as follow:
\begin{itemize}[label=$\sqbullet$]
     \item Full low-level C implementation of the ADMM-LSMR method. 
     \item Implementation of fixed-point arithmetic with four different rounding methods.
     \item Implementation of 16-bit and 32-bit fixed-point variants of LSMR method to be more hardware efficient.
     \item Primary design and implementation of both CPU (host) side and device (FPGA) side of an OpenCL program and applying detailed amendments to emulate the program successfully.
     \item Deployment of the design on Intel Arria® 10 GX FPGA on Intel DevCloud stack which required another set of design amendments to achieve a successful deployment.
     \item Applying multiple stages of optimisation to increase speed and maximise FPGA board utilisation.
     \item Conducting experiments to assess the accuracy and efficiency of the final hardware-accelerated program.
 \end{itemize}
 \section{Outline}
 In the background chapter of this report, first, we briefly described neural networks, ADMM optimisation method and LSMR algorithm. Next, we explored common approaches in hardware acceleration of neural networks followed by a description of FPGAs and their structures. Then, usage of FPGAs in neural network training acceleration is explored. Finally, an overview of the employed technologies, including OpenCL and Intel FPGA development stack is provided.\vspace{8pt}\\
 In chapter \ref{ch3}, software and algorithmic aspects of the implementation are described. First, the ADMM-LSMR algorithm \cite{iso} is explained, and it is demonstrated how and why the method is a perfect candidate for hardware acceleration. Later, details of the low-level implementation of the method, fixed-point arithmetic and their combination are described.\vspace{8pt}\\
 In chapter \ref{ch4}, the route taken to convert the C implementation to an OpenCL accelerated program and finally a fully working FPGA deployment is explained. A separate section is dedicated to applied optimisation techniques and their outcome.\vspace{8pt}\\
 Chapter \ref{ch5} contains the results of experiments conducted to assess the accuracy and time efficiency of the implemented method. \vspace{8pt}\\
 Finally, the report is closed on chapter \ref{ch6}, providing the conclusion and potential areas to be improved. 
 
\chapter{Background}
\label{ch2}
\section{Artificial Neural Networks}
\label{Background-ANN}

The main goal of the neural networks is to find a function $f$ that best approximates some target function $f^*$. This goal is achieved through a process called \textit{training}, where the network learns a set of parameters from the input data. Using the learned parameters to predict the output of a new data is called \textit{inference}.\vspace{8pt}\\
Two key components of a neural network are neurons and layers. Layers are a collection of neurons, and the network is composed by connecting these layers. Different approaches for connecting layers together leads to different types of neural networks like feed-forward neural networks, convolutional neural networks and recurrent neural networks. Each of these networks is suitable for a specific set of applications \cite{Deep}.
\begin{gather}
    \label{NNopt}
    \min_{W} \ell(f(x_0,W), y) 
\end{gather}
The optimisation problem of a neural network can be written as equation \ref{NNopt}, where $\ell$ is a loss function, and $W$ is the learnable parameter. As we mentioned, the goal is to learn $W$ such that we can minimise the difference between the output of $f$ given the input $x_0$ and the actual output $y$.

\subsection{Feed-forward Neural Networks}
\label{FNN}
\begin{figure}[tb]
\centering
\includegraphics[width = 0.9\hsize]{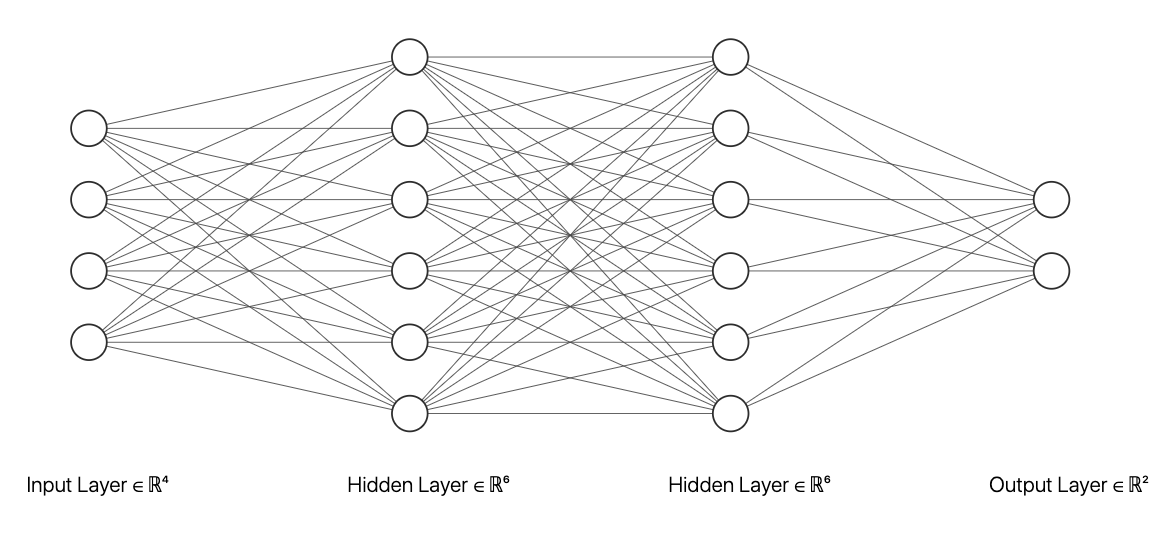}
\caption{A simple feed-forward neural network }
\label{fig:fnn}
\end{figure}
In a feed-forward neural network, all of the neurons of a layer are connected to all of the neurons of the next layer as it is shown in figure \ref{fig:fnn}. This type of neural network is called feed-forward as the information always flows forward in them. It has been proposed that a feed-forward neural network with one hidden layer can approximate any continuous function, but the required hidden size may be significantly large and that would make the process of learning impossible \cite{Deep}.
Feed-forward neural networks are suitable for unstructured data for example when the data is not time-dependent or sequential. \vspace{8pt}\\
Following statements hold for a three-layer feed-forward neural network: 
\begin{gather*}
    \text{Input data }\hspace{5pt}  x_{0} \in {\rm I\!R}^{D*N}  \\
    W_1 \in {\rm I\!R}^{HS*D} \\  z_1= W_1x_0 ,\hspace{5pt} z_1\in {\rm I\!R}^{HS*N}  \\
     \text{Input of hidden layer }\hspace{5pt} x_1 = h_1(z_1) \in {\rm I\!R}^{HS*N}  \\
     W_2 \in {\rm I\!R}^{HS*HS}
      \\
     z_2= W_2x_1 ,\hspace{5pt} z_2\in {\rm I\!R}^{HS*N}  \\
    \text{Input of last layer}\hspace{5pt} x_2 = h_2(z_2) \in {\rm I\!R}^{HS*N} \\ W_3 \in {\rm I\!R}^{OS*HS}
    \\
    \text{Output}\hspace{5pt} z_3= W_3x_2 ,\hspace{5pt} z_3\in {\rm I\!R}^{OS*N} 
\end{gather*} 
Where we have $N$ training samples, $D$ features, $HS$ number of neurons in the hidden layers and $OS$ is the dimensions of output. $h_l$ is the activation function of layer l. This notation is used in this report. 
\newpage
\subsection{Gradient-Based Methods}
\label{Background:gradientbased}
Gradient-based methods \cite{ruder2016overview} are the most common optimisers which are used alongside back-propagation \cite{rumelhart1986learning} to solve the optimisation problem of neural networks. These iterative algorithms use the first-order derivative of the objective function to move towards the optimal solution.\vspace{8pt}\\ 
Many variants of gradient-based methods have been proposed. Such as SGD \cite{bottou1991stochastic} \cite{bottou2012stochastic}, AdaDelta \cite{zeiler2012adadelta}, AdaGrad \cite{duchi2011adaptive}, Nadam and Adam \cite{kingma2014adam} which is the most popular and most commonly used. In general, it is observed that these methods suffer from several limitations, such as:
    \subsubsection{Vanishing and Exploding Gradient}
    This problem which occurs as a result of repeated matrix multiplication, is known as one of the fundamental limitations of the gradient-based methods.
    Multiplying small values of gradient several times results in a very small value that can slow down or even stop the training process. This defect is called vanishing gradient. 
    On the other hand, exploding gradient happens as a consequence of multiplying big values of gradient multiple times. This can make the learning process extremely unstable. In Recurrent Neural Networks, this becomes even more crucial \cite{bengio1994learning}. Proposed methods to reduce the effect of this problem include changing the architecture to Long Short Term Memory networks \cite{hochreiter1997long} and using Rectified Linear Units \cite{nair2010rectified} for activation function which helps with vanishing gradient and also clipping gradients to mitigate exploding gradients.

    \subsubsection{Sequential Dependency}
    Gradient-based methods are sequential by nature. In these methods the gradient computation of a batch can only start after the computation of the previous batch has been completed and weight updates take place when gradients of a batch are available. This characteristic makes the gradient-based methods not a suitable candidate for parallel implementation. This is even a more critical issue in FPGA implementation of neural networks when the parallelisation involves hardware pipelines and an algorithm with fewer dependencies and the ability to be pipelined is desired. 
    
    \subsubsection{Converging to Local Minima or Saddle Points}
    Generally, the optimisation problem of neural networks is non-convex. Therefore, converging to local minima or saddle points is another concern. Also, it has been discussed that in higher dimensions, saddle points can cause more crucial issues \cite{dauphin2014identifying}.
    
    \subsubsection{Sensitivity to Ill-conditioning}
    Another common issue in training neural networks using gradient-based methods is ill-conditioning. When the Hessian of the objective function is ill-conditioned, it can drastically affect the convergence rate of gradient-based methods and slow down the training process \cite{chong1996chong}.


\section{Alternating Direction Method of Multipliers}
\label{Background-ADMM}
         

Recently, Alternating Direction Method of Multipliers (ADMM) \cite{gabay1976dual} has been used as an optimisation method in a wide variety of applications \cite{lin2013design} including machine learning and neural networks \cite{kiaee2016alternating}. This powerful iterative optimisation method breaks down the objective function into smaller pieces that can be solved easier \cite{boyd2011distributed}.\vspace{8pt}\\ 
This method can be applied in parallel, and this characteristic makes it a suitable replacement for gradient-based methods in the optimisation problem of neural networks. 
The inherent parallelism of ADMM also makes it a good option for hardware implementation. However, as we discussed in \cite{iso} this method includes matrix inversion, which is not a hardware friendly operation. ADMM is a combination of dual decomposition and method of multipliers. We have explored the mathematical details of these methods in \cite{iso}.

\section{LSMR}
\label{Background-LSMR}
LSMR \cite{fong2011lsmr} is an iterative method that solves linear systems and least-square problems like \ref{linearsystem} where $A \in {\rm I\!R}^{m * n}$, $b \in {\rm I\!R}^{m}$ and $x \in {\rm I\!R}^{n}$. This method is based on the Golub-Kahan bidiagonalization process \cite{golub1965calculating}, which is shown in pseudo-code \ref{alg:GKB}. Other iterative least-square solvers also exist which are based on Golub-Kahan bidiagonalization process \cite{paige1982lsqr} \cite{estrin2019lslq}. The difference between these methods is usually their early stopping criterion. These algorithms are sequential in principle, but it is worth mentioning that they can be parallelized for solving problems where $X$ and $B$ are matrices as each column can be processed separately. This characteristic is desirable for hardware implementation since it allows pipelining and parallelisation.
\begin{gather}
\label{linearsystem}
    Ax  = b \\ \nonumber
    \min_x ||Ax -b ||_2
\end{gather}

\begin{algorithm}
\SetAlgoLined
\KwIn{$A \in {\rm I\!R}^{m * n} , b \in {\rm I\!R}^{m}$ }
    $\beta_1 \leftarrow ||b||_2 $\\ 
    $u_1 \leftarrow b/\beta_1$ \\
    $\alpha_1 \leftarrow ||A^T u_1||_2$ \\ 
    $v_1 \leftarrow A^T u_1/\alpha_1 $\\
 \For{$k =1,2,...$}
 {
     $\beta_{k+1} \leftarrow ||A v_k - \alpha_k u_k||_2$ \\ 
     $u_{k+1} \leftarrow (A v_k - \alpha_k u_k )/ \beta_{k+1}$\\
     $\alpha_{k+1} \leftarrow ||A^T u_{k+1} - \beta_{k+1} v_k||_2$ \\ 
     $v_{k+1} \leftarrow (A^T u_{k+1} - \beta_{k+1} v_k )/ \alpha_{k+1}$
 }

 \caption{Golub-Kahan Bidiagonalization Process \cite{estrin2019lslq}}
\label{alg:GKB}
\end{algorithm}


\section{ADMM for Neural Networks }
\label{Background-ADMMNN}

The implemented ADMM-LSMR method in \cite{iso} and this work, is based one an ADMM-based training method proposed in \textit{"Training Neural Networks Without Gradients: A Scalable ADMM Approach"} \cite{taylor2016training}. Here, their work is briefly discussed using the notation provided in section \ref{FNN}.\vspace{8pt}\\
To utilise ADMM for solving the optimisation problem of neural networks, the key idea proposed in \cite{taylor2016training} is to use a variable called pre-activation $z_l$ for each layer $l$. This will enable us to decouple the weights from the activation function and rewrite the optimisation problem of an $L$ layer neural network to the following:
\begin{gather}
    \label{admmNN}
    \min_{W_l , x_l,z_l} \ell(z_L, y) 
    \\ 
    \text{subject to  }\hspace{5pt} z_l = W_lx_{l-1}, \text{ for } l = 1,2,...L \nonumber
    \\   x_l = h_l(z_l), \text{ for }l = 1,2,...L-1\nonumber
\end{gather}
The augmented Lagrangian of \ref{admmNN} can be written as:
\begin{gather}
  \ell(z_L, y)  + \beta_L||z_L - W_Lx_{L-1}||_2^2  \\ \nonumber  + \sum_{l=1}^{L-1}[\gamma_l||x_l - h_l(z_l)||_2^2 + \beta_l||z_l - W_lx_{l-1}||_2^2] \\ \nonumber + 
  \sum_{l=1}^{L-1} \lambda_l^T(z_l - W_lx_{l-1}) + \delta_l^T(x_l - h_l(z_l)) \\ \nonumber + 
  \lambda_L^T(z_L - W_Lx_{L-1}) + \delta_L^T(x_L - h_L(z_L)) 
\end{gather}
Where $\lambda_l$ and $\delta_l$ are vectors of Lagrangian multipliers and $\gamma_l$ and $\beta_l$ are penalty parameters.
The proposed method in \cite{taylor2016training} uses just one Lagrangian multiplier vector since they observed that by applying the classic ADMM where each constant has its own Lagrangian vector, the method would be unstable. This results in \ref{ADMML} where the only Lagrangian multiplier vector is $\lambda$

\begin{gather}
    \label{ADMML}
  \ell(z_L, y)  + \beta_L||z_L - W_Lx_{L-1}||_2^2  \\ \nonumber  + \sum_{l=1}^{L-1}[\gamma_l||x_l - h_l(z_l)||_2^2 + \beta_l||z_l - W_lx_{l-1}||_2^2] \\ \nonumber + 
  \lambda^T(z_L - W_Lx_{L-1})
\end{gather}
Pseudo-code of their proposed method is provided in algorithm \ref{alg:ADMM-NN}. In this method, variables are updated one at a time while the others are fixed. Minimisation steps of this algorithm are explained in the following sections.

\begin{algorithm}
\SetAlgoLined
 \While{not converged}{
  \For { $l =1,2,... L-1$} {
  $W_l \leftarrow z_lx_{l-1}^\dagger$ \\ 
  $x_l  \leftarrow ( \gamma_l I +\beta_{l+1}W_{l+1}^T W_{l+1}) ^{-1}(\gamma_lh_l(z_l) + \beta_{l+1}W_{l+1}^T z_{l+1}) $
  \\
  $z_l \leftarrow \argmin_z {\gamma_l||x_l - h_l(z_l)||_2^2 + \beta_l||z_l - W_lx_{l-1}||_2^2 }$

  }
  $W_L \leftarrow z_Lx_{L-1}^\dagger$ \\ 
  $z_L \leftarrow \argmin_z { \ell(z_L, y)  + \beta_L||z_L - W_Lx_{L-1}||_2^2 + \lambda^T(z_L - W_Lx_{L-1})}$\\
  $\lambda \leftarrow \lambda + \beta_L(z_L - W_Lx_{L-1})$

 }
\caption{ADMM for Neural Networks \cite{taylor2016training}}
\label{alg:ADMM-NN}
\end{algorithm}
\subsubsection{Weight Update}
Solution of minimising \ref{ADMML} with respect to $W_l$ can be written as \ref{weightupdate} where $x_{l-1}^\dagger$ is the pseudo-inverse of the matrix $x_{l-1}$.
\begin{gather}
\label{weightupdate}
W_l \leftarrow z_lx_{l-1}^\dagger
\end{gather}
\subsubsection{Activation Update}
$x_l$ is updated using the equation \ref{activationupdate} in each in each step. Details of how this equation is derived of are discussed in \cite{iso}.
\begin{gather}
    \label{activationupdate}
    x_l  \leftarrow ( \gamma_l +\beta_{l+1}W_{l+1}^T W_{l+1}) ^{-1}(\gamma_lh_l(z_l) + \beta_{l+1}W_{l+1}^T z_{l+1}) 
\end{gather}
\subsubsection{Output Update}


The new value of $z_L$ is calculated using the optimisation problem \ref{outputupdate}. This optimisation problem is non-convex and non-quadratic because of the activation function $h$. However, it can be solved easily in closed form when $h$ is piece-wise linear since the activation function works element-wise on its inputs.

\begin{gather}
\label{outputupdate}
    \argmin_z {\gamma_l||x_l - h_l(z_l)||_2^2 + \beta_l||z_l - W_lx_{l-1}||_2^2 }
\end{gather}
\subsubsection{Lagrangian Multiplier Update}
The Lagrangian multiplier is updated using the following equation:
\begin{gather}
      \lambda \leftarrow \lambda + \beta_L(z_L - W_Lx_{L-1})
\end{gather}

\newpage
\section{Hardware for Neural Networks}
\label{Background-HWNN}
As the amount of available data and also the complexity of neural networks are increasing, computation and storage cost of these models are growing rapidly. In some cases, these requirements have made the use of large neural networks impossible, especially in applications where low power consumption or small latency is critical. As a result, choosing and designing efficient computing platforms for neural network applications is becoming more critical than ever \cite{guo2017survey} \cite{sze2017hardware}.\vspace{8pt}\\
Training or inference of neural networks on general-purpose CPUs with von Neumann architecture is inefficient since a significant amount of MAC operations are involved in these processes. CPUs neither have high performance in this area nor low power consumption and are not suitable for either cloud or mobile applications of neural networks. Also, breakdown of Dennard scaling, failure to increase the clock frequency, and the low rate of data transfer between CPU and memory, known as von Neumann bottleneck, have made the use of custom hardware architectures for neural networks more interesting.\vspace{8pt}\\ 
GPUs have a higher arithmetic density compared to CPUs. As a result, nowadays, neural networks are usually trained on GPUs which have very high power consumption. The need for low-power and efficient platform for training neural networks has led to a significant increase in research in using custom hardware architecture for neural networks in the last decades \cite{farabet2011large} \cite{chen2014dadiannao} \cite{esser2015backpropagation}.
\subsection{Motivation}
One of the main reasons for interest in the use of custom hardware is exploiting the inherent parallelism of neural networks. Also, as we mentioned, von Neumann bottleneck and breakdown of Dennard scaling result in limitations for CPUs and highlight the need for more efficient hardware platforms. Computations of neural network models usually take place in the cloud, but, there are some applications where local embedded processing is preferred because of privacy. In these cases, small footprint and low power consumption become more important. These two factors are also critical in wearable or implantable medical devices. Other important motivations are increasing the speed of computation and decreasing the latency, which is always desirable and also is critical in applications like autonomous vehicles and robotics. It has been observed that with custom chips, we can achieve much faster neural networks compared to von Neumann architectures \cite{sze2017hardware} \cite{schuman2017survey}.\vspace{8pt}\\
Considering the above, the main motivations behind the growing interest in hardware implementations of neural networks can be summarised as:
\begin{itemize}[label=$\sqbullet$]
    \item Parallelism
    \item CPU limitations
    \item Low power consumption
    \item Small footprint
    \item High Speed 
\end{itemize}
\subsection{Approaches}
Key factors that should be considered when implementing hardware-based neural networks are the following \cite{sze2017hardware}: 
\begin{itemize}[label=$\sqbullet$]
    \item Accuracy: A common measure that demonstrates how performant a neural network is.
    Accuracy of neural networks should not be compromised for the hardware implementation.
    \item Power consumption: The energy consumed by the platform. Data movements usually cost more energy than computation.
    \item Throughput/latency: How much data can be processed at a time and how fast can the network respond to queries. Latency is more important in inference. 
    \item Cost: Is determined by many factors. If a small number of chips are needed, ASIC design costs much more than FPGA. Also, the complexity of the circuit and the amount of memory required on the chip have a direct impact on the cost.
\end{itemize}

Various techniques have been developed to maintain a trade-off between these factors while making neural networks more hardware-compatible. These techniques usually aim to reduce data movement, computation and required storage on the chip, while maintaining the accuracy. 
\newpage
Hardware implementation of neural networks can be split into three major categories \cite{schuman2017survey} \cite{girau2006fpna}:
\begin{itemize}[label=$\sqbullet$]
    \item Analog: ASIC and FPAA (Field Programmable Analog Arrays).
    \item Digital: ASIC and FPGA.
    \item Mixed Analog/Digital systems.
\end{itemize}
Analog ASIC designs are fast and dense with low power consumption, but, they are expensive and lack flexibility. In general analog designs are noisier than digital designs, and they may suffer from problems such as not being precise and robust along with data storage problems. Another problem with FPAAs is that currently there are very few FPAA manufacturers and their on-chip resources, which are critical in neural network implementations, are much less than FPGAs. Digital ASIC designs provide more accuracy and robustness compared to analog ASIC, but again they are very expensive and not flexible with a very time-consuming and challenging development process. There is also ongoing research to implement mixed analog/digital circuits for networks. These systems have the overhead of ADC and DAC conversion \cite{sze2017hardware} \cite{girau2006fpna}.\vspace{8pt}\\
FPGAs usually are less performant comparing to ASIC designs in term of area, power and speed. On the other hand, they have a faster design process and are less expensive. These reconfigurable platforms also benefit from increased processing density (greater performance per unit of silicon area) compared to general-purpose processors, and they can have better cost:performance ratio compared to both ASIC and general purpose-processors. In addition, FPGAs have the advantage of being reconfigurable, which means that they are flexible and can be programmed to be used on different neural networks on-demand \cite{omondi2006fpga} \cite{liu2009survey} \cite{moussa2006arithmetic}.

\section{Field-programmable Gate Arrays} 
\label{Background-FPGA}
\begin{figure}[tb]
\centering
\includegraphics[width = 1.0\hsize]{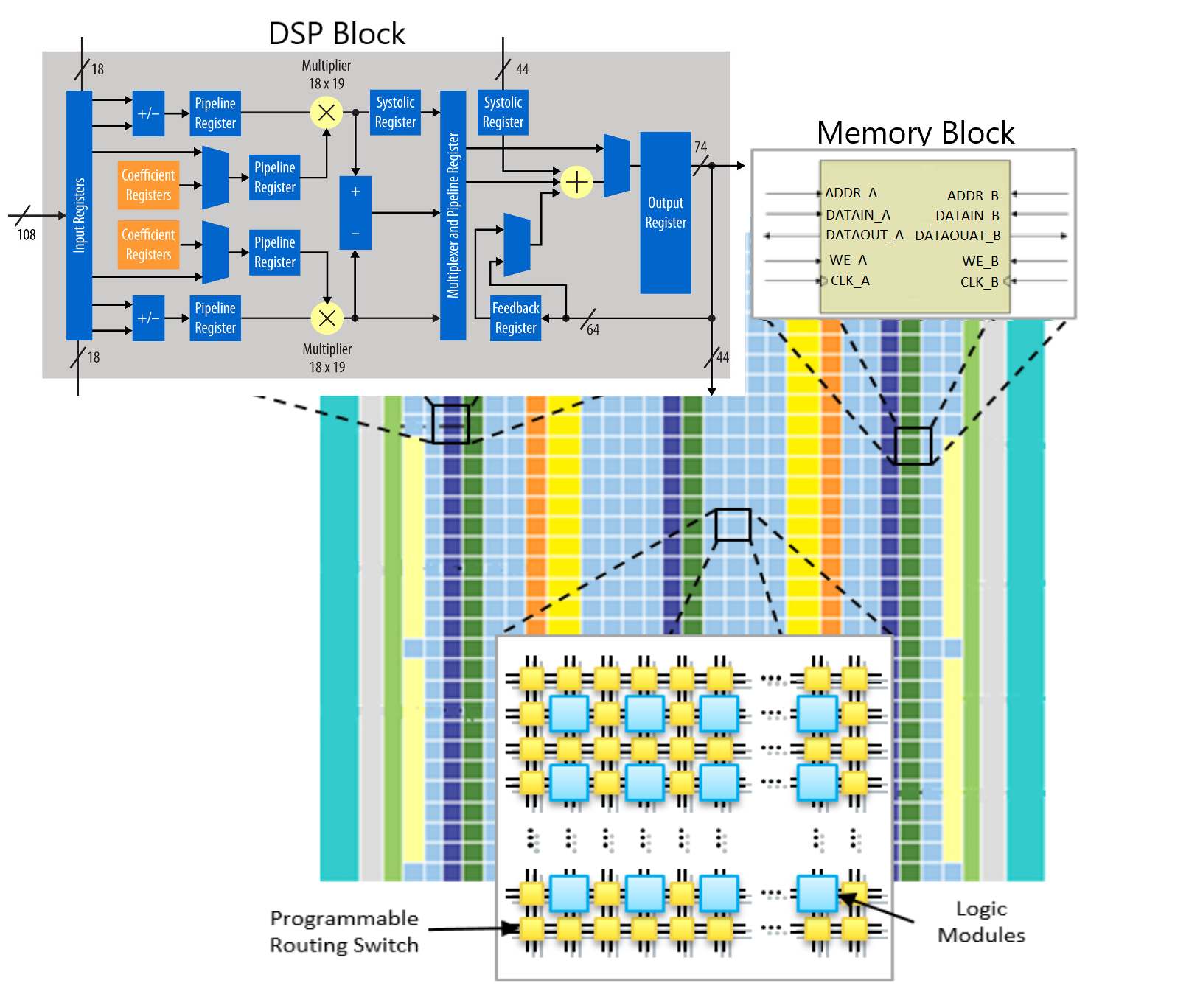}
\caption{FPGA Architecture \cite{bestpracticeguide}}
\label{fig:FPGAarchitecture}
\end{figure}

\begin{figure}[tb]
\centering
\includegraphics[width = 0.6\hsize]{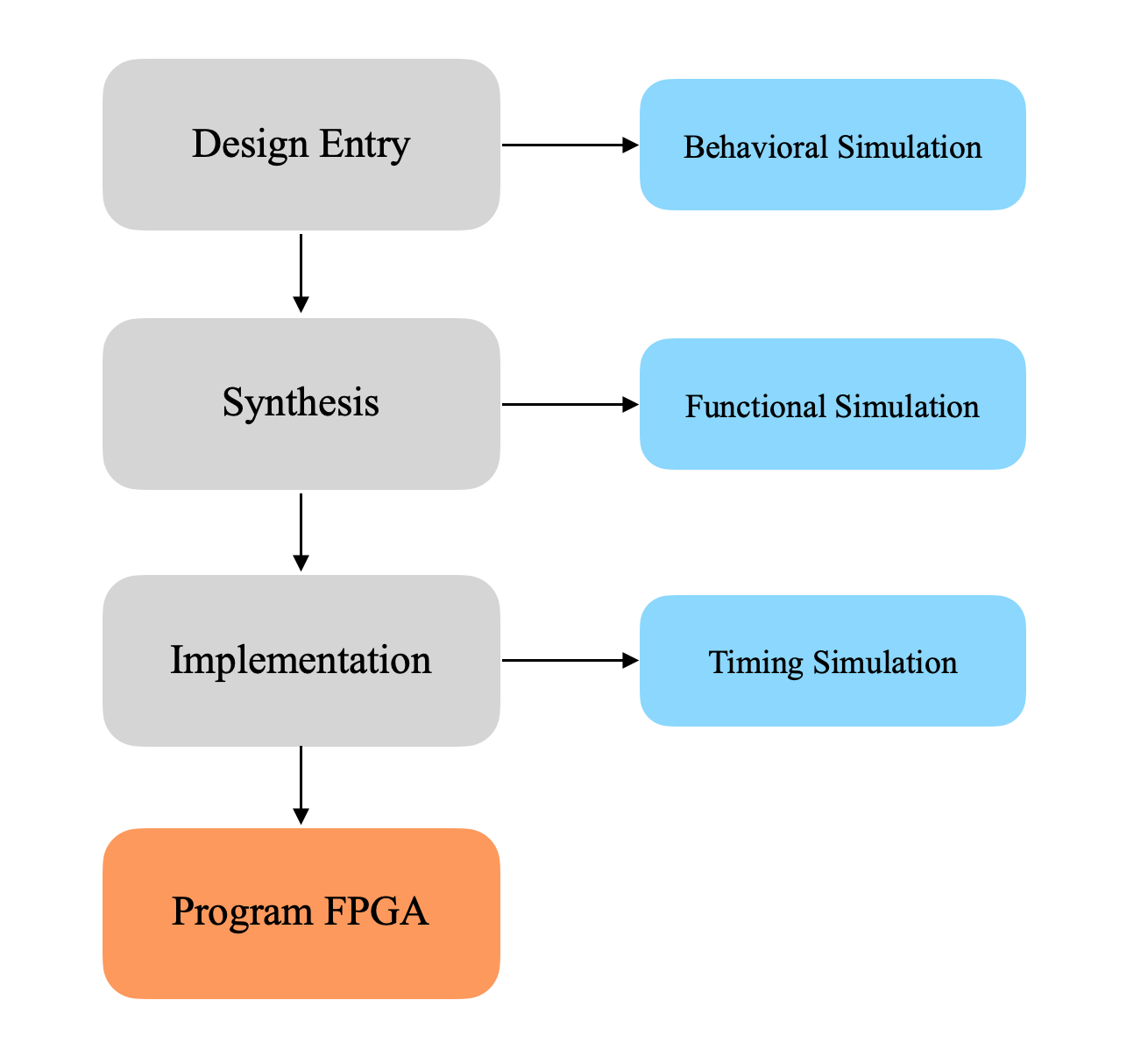}
\caption{FPGA design flow}
\label{fig:FPGAflow}
\end{figure}

A field-programmable gate array or FPGA is a semiconductor integrated circuit (IC) which is made of small computation units, usually called logic blocks, connecting together with programmable interconnections. FPGAs can execute an infinite variety of functions as they can be configured over and over again. Configuring FPGAs is actually programming their logic blocks in a way that their output(s) becomes a specific function of their inputs. Because of this capability, we are able to build custom data-paths and program the dataflow directly into the hardware.\vspace{8pt}\\
A simple view of an FPGA architecture is shown in figure \ref{fig:FPGAarchitecture}. Interconnections have the task of connecting the logic blocks and making the flow of signals inside the chip possible. A logic block usually consists of lookup tables (LUT), flip flops (FF) and multiplexers. The primary structure of FPGA is a two-dimensional array of logic blocks, interconnections and I/O blocks. These days FPGAs usually include on-die processors, RAM blocks, digital signal processors (DSP) and embedded multipliers.\vspace{8pt}\\
The main advantages of FPGAs can be enumerated as following:

\begin{itemize}[label=$\sqbullet$]
    \item They are programmable, and their functionality can be changed by downloading a new configuration file into the device. FPGAs can be considered as platforms that can implement a custom instruction set for a target application. This is while multiple instructions must be combined to perform the same operations in CPUs, DSPs or GPUs.
    \item One of the main advantages of FPGAs is their support for pipeline implementations. In FPGAs, the parallelism is not necessarily achieved by replication of compute units. The pipelining approach allows parallelism while maximising hardware utilisation \cite{bestpracticeguide}.
    \item Besides being re-programmable, FPGA design process is considerably faster and easier than ASIC design. The development cost is also significantly lower compared to ASIC.
\end{itemize}
These features have made FPGAs very popular over the past decades and they are employed in a wide range of applications like speech recognition, image processing, video compression, ASIC prototyping and medical applications. \vspace{8pt}\\
The FPGA design flow is shown in figure \ref{fig:FPGAflow}. Each step is explored in the following paragraphs. \vspace{8pt}\\
Design entry is performed by using schematic or a hardware description language (HDL). By using schematic, the designer has to design the low-level hardware. As a result, this technique can be used only in small projects while HDLs such as Verilog and VHDL can be used for more complex systems and make the design process faster. Recently, this step can also be done using higher-level programming languages like C and let the C-to-FPGA compilers translate the C code into HDL. Such translation is performed when developing FPGA accelerated programmes using OpenCL. By using higher-level languages, the designer has less control over the FPGA resources and may not be able to utilise all of the available hardware resources compared to HDL designs, but, the design process will be less time-consuming. \vspace{8pt}\\
In the synthesis step, the design is translated into a circuit using a netlist. A netlist lists the required logic elements and interconnections. First, a syntax check is applied and then an optimisation process is performed in order to eliminate redundant logic and reduce the size of the design. Next, the details of the design are planned, and the timing properties are estimated.\vspace{8pt}\\
The layout of the design is determined in the implementation step. In this stage, the design is mapped into logic blocks of the FPGA, and then the IO blocks and logic blocks are connected.\vspace{8pt}\\
In the last step, the mapped and routed design is loaded into the FPGA using a generated bitstream file.\vspace{8pt}\\
In order to test the design, at the end of each step, a simulation can be performed.
Behavioural simulation is performed before synthesis to check the functionality of the design. Functional simulation or post-synthesis simulation is a netlist level simulation which ignores the timing.
In timing simulation, wiring and delays are also taken into account. This simulation usually is more time consuming but is the most accurate one \cite{xfpga} \cite{ifpga} \cite{hfpga}.

\section{FPGA for Neural Networks}
\label{Background-FPGANN}
\begin{figure}[tb]
\centering
\includegraphics[width = 0.6\hsize]{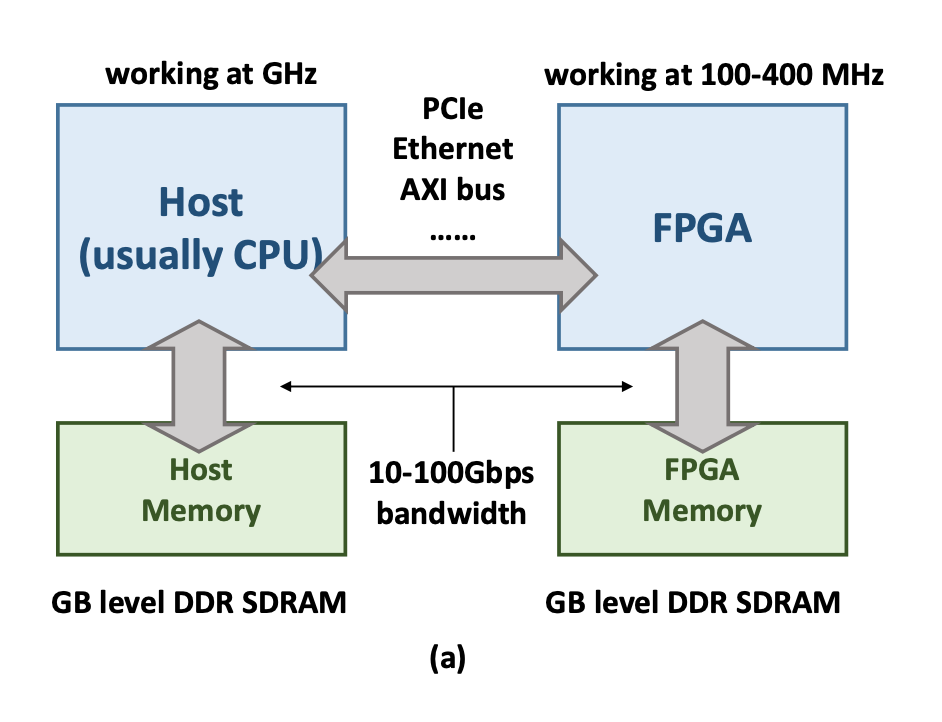}
\caption{A typical FPGA-based Neural Network \cite{guo2017survey}}
\label{fig:FPGANN}
\end{figure}

FPGAs have a parallel architecture which makes them suitable for massive convolution, MAC, and other essential matrix operations in training neural networks \cite{hao2017general}. They also benefit from flexibility in design and short development process like software, while having the performance closer to ASIC designs \cite{moussa2006arithmetic}.\vspace{8pt}\\
An FPGA-based neural network system consists of two parts: CPU part (host) and FPGA part (device). These two parts are usually connected with PCIe connections. FPGAs generally have on-chip SRAM (Static Random Access Memory) which are usually not enough for storing neural network parameters, and we have to use off-chip memory. The performance of the system is usually bounded by bandwidth and power consumption of this external memory. An abstract structure of a typical FPGA implementation of a neural network is illustrated in figure \ref{fig:FPGANN}. The role of the host is to monitor the FPGA and issue commands to it. The FPGA usually has a controller which is responsible for communicating with the host and also issuing signals for other modules in FPGA. This controller can be a finite state machine or a decoder \cite{guo2017survey}.\vspace{8pt}\\
In general, two main factors limit the performance of an FPGA-based neural network: 
\begin{itemize}[label=$\sqbullet$]
    \item On-chip resources.
    \item Off-chip memory bandwidth.
\end{itemize}
The main proposed ideas in order to make neural-networks more suitable for FPGA-implementation fall into three categories \cite{guo2017survey} \cite{sze2017hardware}: 
\begin{itemize}[label=$\sqbullet$]
    \item Reduce precision.  
    \item Sparsity 
    \item Compression
\end{itemize}
In the following sections, first precision reduction techniques are explained in details as this approach is employed in our implementation. Next, an overview of the other two approaches is given. There are also other techniques that do not fall into these categories. The common challenge in all the existing approaches is to make an optimal trade-off between accuracy and hardware speed and energy. For example, by reducing the size of each computation unit, we would be able to place multiple replicas of the units on the FPGA and increase the throughput. One possible way to reduce the compute unite size could be using fixed-point arithmetic which results in sacrificing the precision. It is also worth mentioning that some implementation details like data access pattern can affect the efficiency of hardware utilisation.

\subsection{Precision Reduction}

In order to meet the computation requirements of training neural networks, domain-specific accelerators which have densely packed floating-point arithmetic units are being utilised. One of the factors that limit the speed of training neural networks is the arithmetic density of hardware platforms. As a result, there is a fair amount of ongoing research to replace the floating-point arithmetic with fixed-point or even using less number of bits which increase the performance density \cite{drumond2018training}.\vspace{8pt}\\
Techniques of narrowing arithmetic are widely used in neural network customised hardware and are not specific to FPGA implementations. For example, NVIDIA TESLA V100 GPU \cite{nvidia2017v100} takes advantage of 16bit-32bit mixed-precision arithmetic and google TPU v2 and v3 \cite{jouppi2017datacenter} use Bfloat 16 which is a 16-bit floating-point representation that has been tailored for training neural networks and has a better performance compared to IEEE half-precision representation in neural network applications.\vspace{8pt}\\
One of the main issues of implementing neural networks on FPGA is selecting the best numerical precision. Single and double-precision floating-point representations decrease the quantisation error because of their high precision at the cost of a significant amount of FPGA resources. For example, in FP32 (Single precision floating-point) 24 bits are dedicated for mantissa, which results in a very high precision that is not needed for our purpose. In \cite{drumond2018training} they trained ResNet 20 \cite{he2016deep} on CIFAR-10 using floating-point representations, and they altered mantissa and exponent width to observe the validation error. Their observations can be summarised as following: 1.Convergence without precision loss, using 8-bit mantissa.
2.Convergence with a small precision loss, using 4-bit mantissa. 3.Divergence using 2-bit mantissa. They also mentioned that the exponent width could not be narrowed as it has a significant impact on the representable range. FP16 (half-precision floating-point) is denser than FP32 but still needs more hardware than fixed-point. In this representation, 11 bits are assigned to mantissa, and the remaining 5 are for the exponent. FP16 suffers from the issue of narrow representable range. On the other hand, fixed-point numbers which have less precision and narrow range, increase the quantisation error while requiring less amount of FPGA resources \cite{drumond2018training}.\vspace{8pt}\\
In \cite{micikevicius2017mixed}, it has been discussed that training with fixed-point or half-precision floating-point has mixed results because of the limited representable range. This fact makes a vital trade-off between hardware resources and precision. The precision affects the neural network accuracy and also the speed of its convergence. However, higher precision is associated with more hardware requirement. The challenge is to find an optimal point and a balance between the required precision and hardware resources.\vspace{8pt}\\
Since there is a limited amount of resources available on FPGAs, and in order to make efficient use of them, we aim to find the minimum viable precision and minimum viable range. This is equivalent to finding the maximum amount of quantisation error that can be tolerated without affecting the accuracy drastically. By using fewer bits for neural network computations, we can also reduce the bandwidth requirement, which is one of the main issues of the FPGA-based implementation of neural networks. \vspace{8pt}\\
To summarise, using fewer bits and simpler representation have the following advantages which make the FPGA implementation of neural networks more feasible: 
\begin{itemize}[label=$\sqbullet$]
    \item Less memory requirement
    \item Less computation cost
    \item Less hardware requirement
    \item Less bandwidth requirement
\end{itemize}

Considering these advantages, using standard floating-point numbers is not the best choice and usually, more area-efficient numeric representations like 16 or 32 bit fixed-point are used. In order to use low-precision computation, we have to quantise the weights and activations of the neural network \cite{gupta2015deep}. 
One of the simplest techniques is to use the nearest fixed-point number representation of each parameter. This method suffers from overflow and underflow because the range of floating-point is highly dynamic and easily exceeds the representable range with fixed-point.
It has been found that the range of parameters of a neural network (weights and activation) is limited in a single layer, but, this range differs when comparing different layers \cite{qiu2016going}. It is also possible to use more bits for the first and last layer and utilise ternary or binary representations for hidden layers \cite{wang2018design}.  \vspace{8pt}\\
 Low-precision computation is widely used in the inference part of neural networks to make the run-time faster, and they can usually reach 32-bit floating-point accuracy  \cite{umuroglu2017finn} \cite{chen2016eyeriss} \cite{ghasemzadeh2018rebnet} \cite{han2015deep} \cite{krishnamoorthi2018quantizing} \cite{choi2018pact} \cite{zhou2017incremental} \cite{anwar2015fixed}. Even binary and ternary representations have been used for inference \cite{hubara2017quantized} \cite{li2016ternary} \cite{zhu2016trained}.
  \vspace{8pt}\\
 On the contrary, using low-precision computation in training neural networks usually has an evident negative effect on accuracy. This is mainly due to the nature of back-propagation and gradient-based methods \cite{fox2019training} \cite{siddhartha2018simultaneous} \cite{liu2017fpga} \cite{courbariaux2016binarized} \cite{wang2018training} \cite{han2017ese} \cite{wu2018training} \cite{courbariaux2014training} \cite{banner2018scalable}. For example, in stochastic gradient descent, which is a common optimiser used in neural networks, many small noisy steps take place for each parameter update. It is obvious that to keep track of these small steps, and for SGD to work at all, high precision is required \cite{courbariaux2015binaryconnect} \cite{hubara2017quantized}. One of the widely used methods to tackle this problem is to use high precision for gradient accumulation and then use lower-precision for other parts of the learning \cite{wang2018training} \cite{wu2018training} \cite{courbariaux2015binaryconnect} \cite{rastegari2016xnor}. Gradient accumulators are frequently updated during training, and the fact that storing them in low-precision adversely affects the accuracy is not desirable \cite{yang2019swalp}. Training neural networks with end-to-end low precision has been done in \cite{zhang2017zipml}, \cite{zhou2016dorefa} and \cite{koster2017flexpoint}. They have used different methods to restrict the range of activations and selecting quantisation points. It is also worth mentioning that in \cite{zhou2016dorefa}, parameters of first and last layers of networks are not quantised.
 \vspace{8pt}\\
 In general, we can conclude that while many benefits are observed in researches on low-precision training, reduction in the accuracy is also reported\cite{de2018high}.
 \subsection{Sparsification}
It is possible to reduce the number of MAC operations by removing some weights of the network. There are different approaches to achieve this, including removing the weights with small absolute value \cite{han2015deep} \cite{nakahara2019fpga}, or values with minimal impact on the output. Another approach is to set the value of some weights to zero in order to remove them. This techniques are beneficial for both computation costs and required storage \cite{guo2017survey} \cite{sze2017hardware}.
\subsection{Compression}
As we mentioned before, data movement and storage are two critical issues in implementing FPGA-based neural networks. Using compression techniques helps with both of these issues. In order to compress parameters of a neural network, both lossless and lossy compression can be utilised. In some cases, codes are assigned to values and a translation table should be used \cite{han2015deep} \cite{han2016eie} \cite{chen2015compressing} \cite{guo2017survey} \cite{sze2017hardware}. Using low-precision for inference of neural networks can also be considered as a type of compression. In \cite{samragh2017customizing}, they proposed a greedy algorithm to encode the parameters of networks considering the platform's memory and the accuracy required for the given task. Also, in \cite{chen2015compressing}, they used a hash function to compress and reduce the size of a trained model.


\section{Fixed-point Arithmetic}
\label{Background-fixedpoint}
As previously stated, fixed-point data types are widely used in FPGA implementation of neural networks. In general, fixed-point numbers can be utilised whenever performance is more critical than precision. \vspace{8pt}\\
Two parameters are associated with a fixed-point data type definition: 
\begin{itemize}[label=$\sqbullet$]
    \item Bit width of representation. We refer to this as word length ($WL$).
    \item Number of fractional bits which determines the position of the binary point. We refer to this as fraction length ($FL$).
\end{itemize}
We will use the notation $fixed \langle WL, FL\rangle$ in this report. We can also calculate the integer length ($IL$), the representable range ($RR$) and the smallest representable positive number ($\epsilon$) as the following:
\begin{gather}
    IL = WL - FL.\\ 
    RR = [ -2^{IL-1} , 2^{IL-1} - 2^{-FL}]\\
 \epsilon = 2^{-FL}
\end{gather}

The $\epsilon$ is a crucial parameter in fixed-point data type and is used frequently in fixed-point arithmetic. 
It is important to highlight the fact that in computer a bit pattern can represent different values in different number systems. When working with fixed-point numbers we should consider both the \textbf{\textit{representation}} and the \textbf{\textit{value}} of a number. In a given $fixed\langle WL, FL\rangle$ we have $FL$ number of fraction bits in \textbf{\textit{representation}} and to interpret the \textbf{\textit{value}} of a given bit pattern we have to do the following:
\begin{enumerate}
    \item Calculate the value of the bit pattern in two's complement format.
    \item Divide the calculated value by $2^{FL}$ (or multiply with $\epsilon$).
\end{enumerate}
Consider the following examples with $fixed\langle 16, 10\rangle$  data type:
\begin{gather}
  \epsilon \text{ \textbf{\textit{representation}}: } 0000 0000 0000 0001 \\ 
    \epsilon \text{ \textbf{\textit{value}} : } 2 ^ {-10}
\end{gather}
\begin{gather}
     \text{ \textbf{\textit{representation}}: } 0101 1100 1000 1001  \\ \nonumber
     \text{ \textbf{\textit{value}} : } 23689 * 2^{-10} = 23.1337890625
\end{gather}
\begin{gather}
     \text{ \textbf{\textit{representation}}: } 1001 0001 1010 0010  \\ \nonumber
     \text{ \textbf{\textit{value}} : } -28254 * 2^{-10} = -27.591796875
\end{gather}

It is also worth mentioning that the two's complement representation can be considered a special case of fixed-point representation with $FL = 0$.\vspace{8pt}\\
As we mentioned earlier, although using fixed-point data types can save a considerable amount of hardware resources and computation, the precision and representable range of a floating-point data type with equivalent word length is significantly higher. 
\begin{table}[H]
\def\arraystretch{1.5}%
\caption{Comparing representable range and smallest positive representable number in floating-point and fixed-point data types}
\begin{center}
\label{tab:RRepsilon}
\begin{tabular}{|l|l|l|}
\hline
\textbf{Data Type}                           & \textbf{Representable Range}                                        & \textbf{\begin{tabular}[c]{@{}l@{}}Smallest Positive \\ Representable Value\end{tabular}} \\ \hline
\textit{ \begin{tabular}[c]{@{}l@{}}32bit Floating-point\\ IEEE 754 Single Precision\end{tabular} }  & $[-3.4 * 10^{38} , +3.4 * 10^{38}]$ & $1.18 * 10^{-38}$     \\ \hline
$fixed\langle32,18\rangle$ &  $[-2 ^{13} , 2 ^{13} - 2 ^{-18}]$                            & $2 ^{-18}$       \\ \hline
\end{tabular}
\end{center}
\end{table}


\subsection{Rounding Methods}   
As the precision of floating-point data types is less than fixed-point numbers, a rounding procedure is required when converting from floating-point to fixed-point.
Different rounding methods can be used for this conversion. The implemented methods in this project are the following:
\begin{itemize}[label=$\sqbullet$]
    \item Downward rounding: Rounds to the nearest representable number which is smaller than the input.
    \item Upward rounding: Rounds to the nearest representable number which is bigger than the input.
    \item Nearest rounding: Rounds to the nearest representable number to the input.
    \item Stochastic rounding: Rounds to the nearest representable number greater or less than the input based on a probability calculated proportionally to the distance.  
\end{itemize}

Given a number $x$ and a fixed-point representation $fixed
\langle WL,FL \rangle$ we define $\lfloor x \rfloor$ as the largest integer multiple of $\epsilon =2^{-FL}$ less than or equal to $x$. The above methods can be mathematically defined as:
\begin{gather}
downwardRound(x) =  \lfloor x \rfloor   
\end{gather}

\begin{gather}
upwardRound(x) =  \lfloor x \rfloor + \epsilon   
\end{gather}

\begin{gather}
    nearestRound(x) =     
    \begin{cases}
       \lfloor x \rfloor  & \text{if   $ \frac{x - \lfloor x \rfloor}{\epsilon} < 0.5$ }\\
       \lfloor x \rfloor + \epsilon & \text{otherwise}\\
    \end{cases} 
\end{gather}

\begin{gather}
    stochasticRound(x) =     
    \begin{cases}
       \lfloor x \rfloor  & \text{with probability $1 - \frac{x - \lfloor x \rfloor}{\epsilon}$ }\\
       \lfloor x \rfloor + \epsilon & \text{with probability $ \frac{x - \lfloor x \rfloor}{\epsilon}$ }\\
    \end{cases} 
    \label{SR}
\end{gather}

Fixed-point quantisation with stochastic rounding has shown promising results in training neural networks with gradient-based methods. However, it has the overhead of pseudo-random number generator compared to nearest rounding \cite{gupta2015deep} \cite{li2017training}. 
\\ 



\section{Development Stack}

\subsection{OpenCL}
\label{Background-opencl}
OpenCL (Open Computing Language) is a free standard for parallel-programming across heterogeneous processing platforms including CPUs, GPUs, DSPs, FPGAs and other processors or hardware accelerators. This framework is used to write portable yet efficient programmes.
\\
The OpenCL specification \cite{khronos2012opencl} uses four models to describe OpenCL concepts:
\begin{itemize}[label=$\sqbullet$]
    \item Platform Model
    \item Memory Model
    \item Execution Model
    \item Programming Model
\end{itemize}

\subsubsection{Platform Model}
\begin{figure}[tb]
\centering
\includegraphics[width = 0.7\hsize]{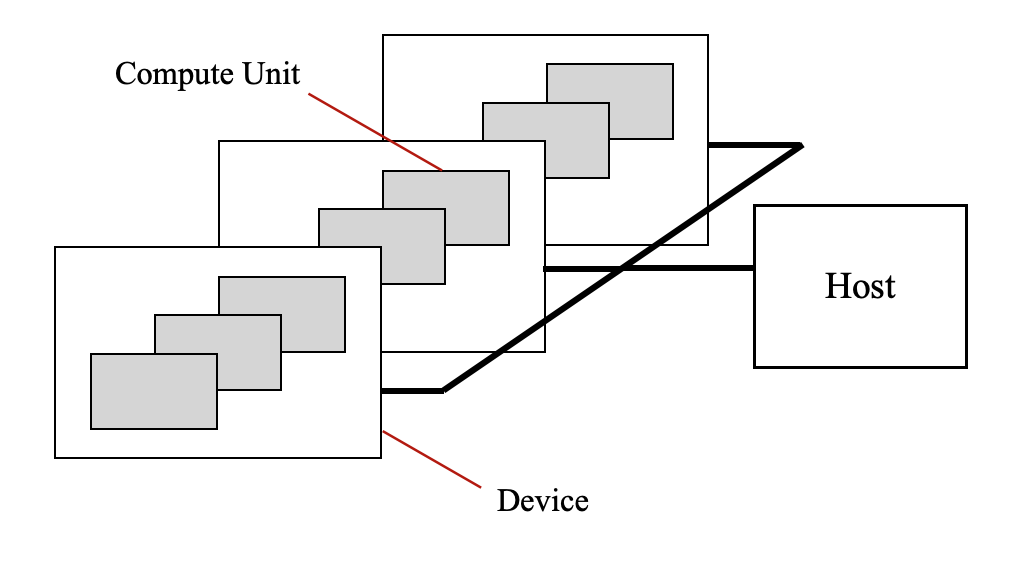}
\caption{OpenCL platform model}
\label{fig:platform}
\end{figure}

The platform model contains a host connected to one or more devices, as shown in figure \ref{fig:platform}. Each OpenCL device consist of one or more compute units. To perform computations on the compute units of a device, relevant commands should be submitted from the host.

\subsubsection{Execution Model}
OpenCL programmes execute in two parts:
\begin{enumerate}
    \item Kernels that execute on devices (CPUs, GPUs, FPGAs and DSPs)
    \item Host program that executes on the host (Usually a general-purpose CPU)
\end{enumerate}
Unit of concurrent execution in OpenCL standard is work-item. Each work-item executes the kernel body. A single iteration of a loop is usually mapped to a work-item. Work-items are divided into work-groups.
 The host program manages the execution of kernels by defining and controlling a context. The context includes the following:
\begin{itemize}[label=$\sqbullet$]
    \item Devices: A list of available devices.
    \item Kernels: Functions that run on devices.
    \item Program Objects: Kernels executable
    \item Memory Objects: Memory objects that are visible to host and devices
\end{itemize}
The host creates one or more command-queues for each device and enqueues different commands to them to manage the execution of kernels. Different queues run independently and concurrently. Commands can be kernel execution commands, memory commands or synchronisation commands. A command queue can accept all the command types and schedules them. Commands in a command queue can execute in-order or out-of-order relative to each other. Whenever a kernel execution or a memory command is submitted to a queue, an event is created. These events can be used by the host program and other commands. Using these events, execution of different commands and their dependencies on each other can be orchestrated and synchronisation points of host program and kernels are managed. 

\subsubsection{Memory Model}

Four different memory regions are defined in OpenCL:
\begin{enumerate}
    \item Global Memory: All work-items in all work-groups have read/write access to this memory. Host can access global memory.
    \item Local Memory: This memory is local to a work-group and all work-items within a work-group have read/write access to this memory.
    \item Constant Memory: A subset of global memory that does not change during the execution of a kernel.
    \item Private Memory: This memory is private to a work-item and is not visible to other work-items.
\end{enumerate}

\subsubsection{Programming Model}
OpenCL supports data-parallel and task-parallel programming models.
By using data-parallelism, we are able to apply a sequence of instruction on different elements of memory. In task parallelism, we execute a kernel using one work-item. We can then enqueue multiple tasks to achieve parallelism.



\subsection{Intel® FPGA SDK for OpenCL}
\label{Background-intelopencl}
\begin{figure}[tb]
\centering
\includegraphics[width = 0.9\hsize]{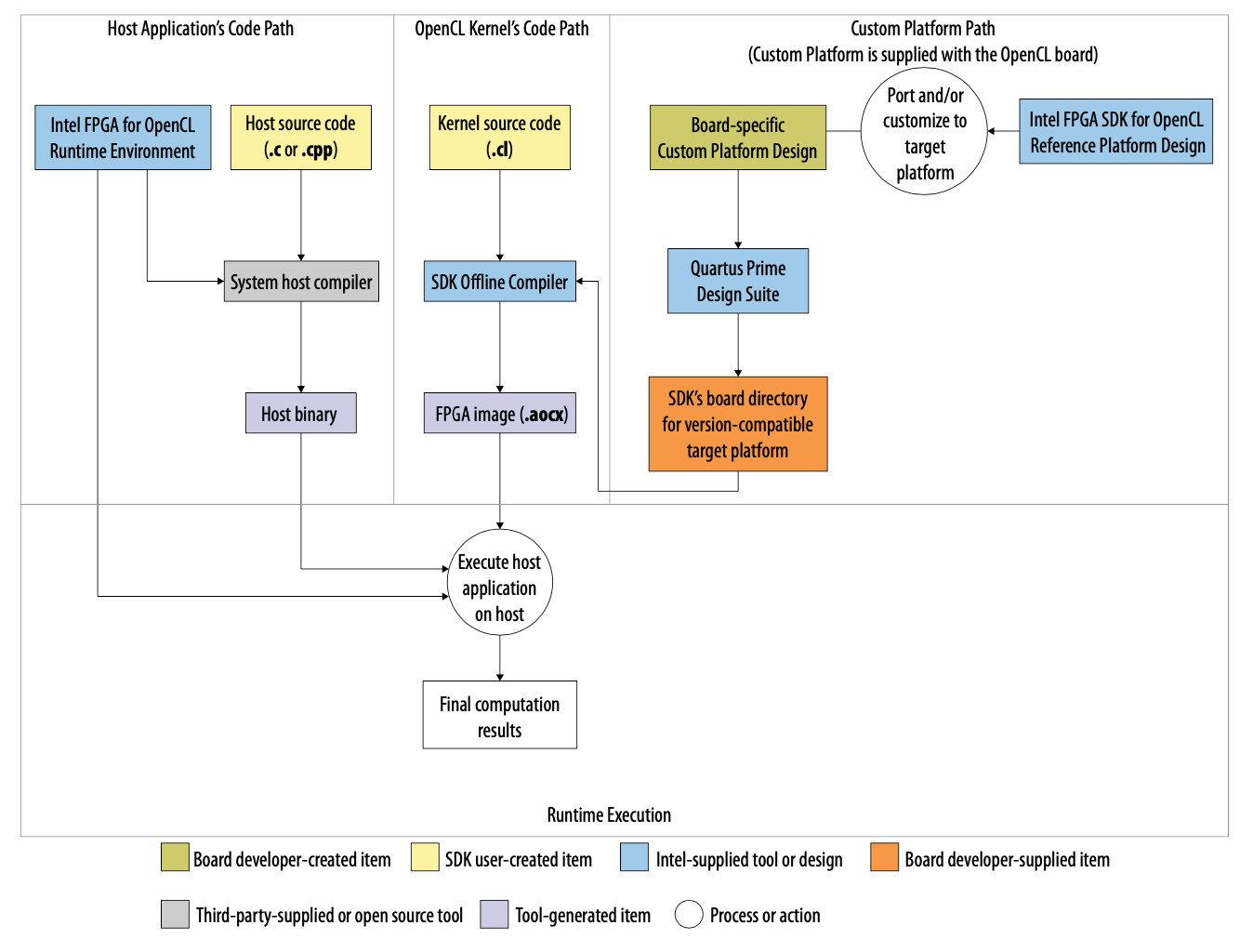}
\caption{Diagram of the Intel FPGA SDK for OpenCL programming model \cite{programmingguide}}
\label{fig:CLmodel}
\end{figure}

Intel FPGA SDK for OpenCL \cite{programmingguide} provides a compiler and a set of tools for building and running OpenCL programmes on Intel FPGA products. Two main components of applications implemented using Intel FPGA SDK for OpenCL are:
\begin{itemize}[label=$\sqbullet$]
    \item Bitstream for programming FPGA
    \item Host program for managing the application flow and FPGA
\end{itemize}
This software development kit includes two compilers. A C++ compiler and an AOC compile. AOC is a specialised offline compiler that compiles C code written for the FPGA (OpenCL kernel) to generate emulator executable or a hardware programming image. The regular C++ compiler generates the executable that runs on the host.\vspace{8pt}\\
As it is shown in figure \ref{fig:CLmodel}, first, the offline compiler compiles OpenCL kernels to an FPGA image file with \textbf{.aocx} extension. This image file is then is used by the host to program the FPGA. The C++ compiler in the host side compiles the host program and links it to the run time libraries of Intel FPGA SDK for OpenCL. The host application, which has the task of programming and executing the hardware image onto the FPGA, is then run by the host.\vspace{8pt}\\
As we explained, in order to program an Intel FPGA the following components should work jointly:
\begin{itemize}[label=$\sqbullet$]
    \item The host compiler
    \item The host application
    \item The offline compiler
    \item The OpenCL kernel(s)
    \item The custom platform
\end{itemize}
When a kernel is compiled, a custom dataflow circuit is generated, and acompute unit is made of different pre-optimised components including load/store units, arithmetic units and flow control units. These components are connected together depending on the dataflow that is implied by the kernel(s).
One of the main advantages of Intel FPGA SDK for OpenCL is that it enables the use of FPGAs without requiring programs written in their specific programming languages such as VHDL or Verilog. Another important feature of Intel FPGA SDK for OpenCL is the detailed report generation in \textit{html} format which contains plenty of information about the resource and area usage alongside performance bottlenecks. We will discuss the information provided in this report more in section \ref{Background-AOCreport}. 

\subsection{Intel AOC Compiler Report}
\label{Background-AOCreport}
Intel recommends using single-work item kernels which are called task instead of NDRange kernels. When a kernel is written as a task, the Intel FPGA SDK FOR OpenCL compiler is able to heavily apply pipeline parallelism to iterations of the loops to achieve high-throughput.\vspace{8pt}\\
The kernel's report.html file gives us kernel analytical data including memory and area usage and also kernel pipeline information. This report includes a summary section alongside three categories of information that can be enumerated as: 
\begin{enumerate}
    \item Summary
    \item Throughput Analysis
    \item Area Analysis 
    \item System Viewers 
\end{enumerate}
We discuss each these sections in more details in the following.

\subsubsection{Summary Report}
Provides an overview of the design, including compile information such as target FPGA family, device and board and AOC version, basic information about kernels including the number of their compute units and their resource usage and achieved fmax of the design.

\subsubsection{Throughput Report}
 The information in this section includes the fmax, bottleneck summary, loop analysis and latency estimation which aids the developer with optimising the kernel. This section is divided into two parts:
 \begin{itemize}[label=$\sqbullet$]
    \item \textbf{Loop Analysis:} Provides useful information for all of the loops, including whether the loops are pipelined or not, and their II (initiation interval). These information helps the developer to maximise the throughput of the designed kernel.\vspace{8pt}\\
    Pipelined loops make efficient use of hardware by keeping more resources occupied and making the process of several data chunks concurrently possible.\vspace{8pt}\\
    \textbf{II or initiation interval} is an important parameter in loop pipelining. When a loop is pipelined, the next iteration will begin before the previous one is finished. \textbf{II} determines the number of clock cycles that are required for launching a new iteration. This is actually the number of clock cycles that is needed to resolve dependencies between iterations of the loop.
     It is obvious that smaller values of II are desirable. In the best case, II value is equal to one, which means that one loop iteration is launched every clock cycle. 
     \item \textbf{Fmax Report:} 
     The scheduled fmax of all the blocks is provided in this section. The maximum frequency at which the output of registers is updated is called the \textbf{fmax}. 
     The duration of the clock cycle is limited by the physical propagation delay of signals between two successive registers which is a function of the complexity of logic of the path. The path with the highest delay limits the speed of the entire circuit and is called the critical path. The \textbf{fmax} is calculated as the inverse of the critical path delay. 
     A high value of fmax means higher performance when there is no other bottleneck and hence is desirable. The AOC compiler tries to optimise the design to achieve the highest possible fmax. When the value of desired fmax and II are not specified in the design, the compiler uses a heuristic to achieve the best fmax/II trade-off.
 \end{itemize}

\subsubsection{Area Report}
This report provides details about the resource utilisation of the kernels, which helps with optimising the kernel to be more area efficient. The resource usage information is available in three levels of hierarchy:
\begin{itemize}[label=$\sqbullet$]
    \item System area: Resources that are utilised by all kernels, including global interconnects and board interface.
    \item Kernel area: Resources that are utilised by each of the kernels of the design, including kernel dispatch logic.
    \item Block area: Resources that are utilised by each of the blocks inside the kernels. Each Block is usually a branch-free section of the code like a loop body.
\end{itemize}

\subsubsection{System Viewers}
This section presents a graphical representation of the generated hardware. This section has three parts:
\begin{itemize}[label=$\sqbullet$]
    \item Graph viewer: Provides graphical report that includes information about sizes of loads and stores, latency and stalls.  
    \item Kernel memory viewer: Demonstrates the AOC compiler interpretation of the data movements of the kernels.
    \item Schedule viewer: Illustrates the scheduled cycle and latency of a group of instructions in the design.
\end{itemize}

\subsection{Intel® FPGA Devcloud}
\label{Background-inteldevcloud}
Intel FPGA Devcloud \cite{inteldevcloud} is an Intel hosted cloud service which provides Intel XEON processors and FPGA acceleration cards for the developers to devise and test their designs. This cloud infrastructure allows users to experiment with their designs' functionality on high-end FPGA accelerator cards. The access to this service will be granted upon request. You will benefit from remote access to Intel servers which are equipped with:
\begin{itemize}[label=$\sqbullet$]
    \item Latest Intel FPGA programmable acceleration cards like Intel Stratix 10 and Intel Arria 10 devices
    \item Intel Core processors 6th to 8th generation 
    \item Intel optimised frameworks and libraries
    \item Software tools needed for FPGA design, development and workload testing, including Intel FPGA SDK for OpenCL.
    
\end{itemize}

\chapter{Software Design}
\label{ch3}



As we mentioned in the previous sections, there are some limitations in gradient-based methods that encouraged us to look for a better solution to solve the optimisation problem of neural networks. 
The proposed method in \cite{iso} which is based on \cite{taylor2016training}, applies ADMM \cite{boyd2011distributed} for training feed-forward neural networks in order to make this process more feasible for hardware implementation. This method implements a simple version of LSMR \cite{fong2011lsmr} as an iterative least-squares method to avoid performing matrix inversion.
As it has been discussed in \cite{iso}, the key characteristics of this method are the following:
\begin{itemize}[label=$\sqbullet$]
    \item Since the method does not use the gradient-based optimisers, their sequential dependency is avoided, and the method is parallel by nature. To be more specific, line 2 and 3 in algorithm \ref{alg:ADMM-LSMR} (\textbf{\textit{weight\_update}} and \textbf{\textit{activation\_update}} procedures) can run in parallel since they don't have any dependencies.
    \item The proposed method had the potential of being combined with fixed-point arithmetic, and since back-propagation is not used in this method, it was expected that the accuracy would not be severely affected.
    \item By using an iterative least-squares method, there is no need to perform matrix inversion, which is the only obstacle for hardware-implementation of ADMM-based training method. In this implementation, LSMR is used to avoid matrix inversion and pseudo-inversion.
    \item LSMR is perfectly suitable for pipeline parallelism as it is an iterative method. This can be observed in algorithm \ref{alg:LSMR}. It also allows independent computation of each column of the result.
\end{itemize}
The pseudo-code of ADMM-LSMR method for training feed-forward neural networks and the implemented LSMR can be seen in the algorithms \ref{alg:ADMM-LSMR} and \ref{alg:LSMR} respectively.

\begin{algorithm}

\SetAlgoLined
\SetKwProg{Fn}{Function}{ } {end}
 \While{not converged}{
  \For { $l =1,2,... L-1$} {
  $
  W_l \leftarrow \textbf{ weight\_update}(z_l, x_{l-1})$\\
  $x_l  \leftarrow \textbf{ activation\_update}(W_{l+1}, z_{l+1}, z_l, \beta, \gamma )$\\
 $ z_l \leftarrow \argmin_z {\gamma_l||x_l - h_l(z_l)||_2^2 + \beta_l||z_l - W_lx_{l-1}||_2^2 }$
  \\
  }
  $W_L \leftarrow \textbf{weight\_update}( z_L, x_{L-1})$\\ 
  $z_L \leftarrow \argmin_z { \ell(z_L, y)  + \beta_L||z_L - W_Lx_{L-1}||_2^2 + \lambda^T(z_L - W_Lx_{L-1})}$\\
 $ \lambda \leftarrow \lambda + \beta_L(z_L - W_Lx_{L-1})$
 
 }

\Fn{weight\_update}
{
\KwIn{$ z_l \in {\rm I\!R}^{m * n} , x_{l-1} \in {\rm I\!R}^{p * n} $}
\KwOut{$W_l \in {\rm I\!R}^{m * p}$}
 \For{$i =1, 2, ..., m $}
 {
  $  W_l^T[:,i] \leftarrow \textbf{LSMR}( x_{l-1}^T, z_l^T[:,i])$\\
 }
}

\Fn{activation\_update}
{
\KwIn{$W_{l+1} \in {\rm I\!R}^{m * n}, z_{l+1} \in {\rm I\!R}^{m * p}, z_l \in {\rm I\!R}^{p * n}, \beta, \gamma$}
\KwOut{$x_l \in {\rm I\!R}^{n * p}$}

   $ part1 \leftarrow \gamma_l I +\beta_{l+1}W_{l+1}^T W_{l+1} $\\
    $part2 \leftarrow \gamma_lh_l(z_l) + \beta_{l+1}W_{l+1}^T z_{l+1}$ \\
 \For{$i =1, 2, ..., m $}
 {
   $ x_l[:,i] \leftarrow \textbf{LSMR}(part1 , part2[:,i])$\\
 }
}

 \caption{ADMM-LSMR for Neural Networks \cite{iso}}
\label{alg:ADMM-LSMR}
\end{algorithm}

\begin{algorithm}
\SetAlgoLined
\SetKwProg{Fn}{Function}{ } {end}
\Fn{LSMR}{
\KwIn{$\mat{A} \in {\rm I\!R}^{m * n} , \vec{b} \in {\rm I\!R}^{m}$ }
\KwOut{$\vec{x} \in {\rm I\!R}^{n}$}
    $\beta_1 \leftarrow ||\vec{b}||_2 ,  \hspace{50pt} \vec{u_1} \leftarrow \vec{b}/\beta_1$\\ 
    $\alpha_1 \leftarrow ||\mat{A^T} \vec{u_1}||_2 , \hspace{25pt} \vec{v_1} \leftarrow \mat{A^T} \vec{u_1}/\alpha_1$ \\ 
    $ \overline{\zeta_1} \leftarrow \alpha_1 * \beta_1 ,\hspace{42pt} \overline{\alpha_1} \leftarrow \alpha_1$ \\
    $\rho_1 \leftarrow \overline{\rho_1} \leftarrow \overline{c_1} \leftarrow 1,\hspace{17pt} \overline{s_1} \leftarrow \zeta_1 \leftarrow 0 $\\
    $\vec{h_1} \leftarrow \vec{v_1}$\\
    $\vec{x} \leftarrow \vec{\overline{h_1}} \leftarrow \vec{0}$\\
    
 \For{$k =1, 2, ..., min(m,n) $}
 {
     $\beta_{k+1} \leftarrow ||\mat{A} \vec{v_k} - \alpha_k \vec{u_k}||_2$ \\ 
     $\vec{u_{k+1}} \leftarrow (\mat{A} \vec{v_k} - \alpha_k \vec{u_k} )/ \beta_{k+1}$\\
     $\alpha_{k+1} \leftarrow ||\mat{A^T} \vec{u_{k+1}} - \beta_{k+1} \vec{v_k}||_2$ \\ 
     $\vec{v_{k+1}} \leftarrow (\mat{A^T} \vec{u_{k+1}} - \beta_{k+1} \vec{v_k} )/ \alpha_{k+1}$\\
     $ c_{k+1}, s_{k+1}, r_{k+1} \leftarrow \textbf{\textit{sym}} (\overline{\alpha_k} , \beta_{k+1})$ \\ 

     $\overline{\alpha_{k+1}} \leftarrow c_{k+1} * \alpha_{k+1}$\\
     $\overline{c_{k+1}}, \overline{s_{k+1}}, \overline{\rho_{k+1}} \leftarrow \textbf{\textit{sym}}(\overline{c_k} * \rho_k, s_{k+1} * \alpha_{k+1}) $\\
     $\zeta_{k+1} \leftarrow \overline{c_{k+1}} * \overline{\zeta_k} ,\hspace{14pt} \overline{\zeta_{k+1}} \leftarrow - \overline{s_{k+1}} * \overline{\zeta_k}$\\

     $\vec{\overline{h_{k+1}}} \leftarrow - (\overline{s_k} * \rho_{k+1} * \rho_{k+1}) / ( \rho_k * \overline{\rho_k} ) \vec{\overline{h_k}} + \vec{h_k}$\\
     $\vec{x_{k+1}} \leftarrow (\zeta_{k+1} / (\rho_{k+1} * \overline{\rho_{k+1}}) \vec{\overline{h_{k+1}}} + \vec{x_k} $ \\
     $ \vec{h_{k+1}} \leftarrow - ((s_{k+1} * \alpha_{k+1}) / \rho_{k+1}) \vec{h_k} + \vec{v}$
 }
}

\Fn{sym}{
\KwIn{ a, b } \KwOut{c, s, r}
\If {abs(b) $>$ abs(a)}
{   $\tau \leftarrow a / b$\\
    $s \leftarrow sign(b) / sqrt(1 + \tau^2)$\\
    $c \leftarrow s * \tau,\hspace{15pt} r \leftarrow b / s$\\
    
}
\Else
{
    $\tau \leftarrow b / a$\\
    $c \leftarrow sign(a) / sqrt(1 + \tau^2)$\\
    $s \leftarrow c * \tau ,   \hspace{15pt}  r \leftarrow a / c $ \\

}
}

 \caption{LSMR }
\label{alg:LSMR}
\end{algorithm}


\definecolor{mGreen}{rgb}{0,0.6,0}
\definecolor{mGray}{rgb}{0.5,0.5,0.5}
\definecolor{mPurple}{rgb}{0.58,0,0.82}
\definecolor{backgroundColour}{rgb}{0.98,0.98,0.99}

\lstdefinestyle{CStyle}{
    backgroundcolor=\color{backgroundColour},   
    commentstyle=\color{mGreen},
    keywordstyle=\color{magenta},
    numberstyle=\tiny\color{mGray},
    stringstyle=\color{mPurple},
    basicstyle=\footnotesize,
    breakatwhitespace=false,         
    breaklines=true,                 
    captionpos=b,                    
    keepspaces=true,                 
    numbersep=5pt,                  
    showspaces=false,                
    showstringspaces=false,
    showtabs=false,                  
    tabsize=2,
    language=C++
}

\section{C Implementation}
\label{achievements-c}
As the first step towards a hardware implementation, a low-level C implementation was required. Each of the sub-procedures of the method including LSMR, \textbf{\textit{weight\_update}}, \textbf{\textit{activation\_update}} ,\textbf{\textit{output\_update}} and \textbf{\textit{lagrangian\_update}}) have been implemented and tested individually and a likewise comparison with Python modules implemented in \cite{iso} was performed for a sanity check.
\subsection{Motivation}
A Python version of the proposed method had been implemented in \cite{iso}. This implementation was heavily using the NumPy library for matrix operations, and the underlying details were out of control of the developer. First of all, C implementation is required for an OpenCL accelerated program as the host program can only be written in C or C++. Secondly, the parts which were aimed to be accelerated also should have been implemented in C because a high-level Python implementation can not be used as a reference to measure the speed up gain of the accelerated version. On the other hand, the device kernel programming language is a derivation of C language and a C implementation which is easier to be tested and modified, can be converted to an OpenCL kernel with minimal effort.
\subsection{Code Structure}
\label{ccode}
This program takes the number of hidden layers in the network and the number of neurons in each layer as input. \vspace{8pt}\\
In order to implement algorithms \ref{alg:ADMM-LSMR}  and \ref{alg:LSMR} in C, we implemented a data structure for storing matrices and also functions to perform primary matrix operations. In this implementation, we used double data type to store the data of matrices.
We defined the matrix data type as a struct:
\begin{lstlisting}[style=CStyle]
typedef struct 
{
    int rows;
    int cols;
    double * data;
} matrix;
\end{lstlisting}
In this struct, we store the number of rows and columns of the matrix and a pointer to where elements of the matrix are stored row-wise in the memory. Numerous matrix operations were required to be implemented for this matrix datatype. There was no particular challenge associated with this implementation apart from dealing with low-level concepts of C language and avoiding memory leaks.
\\

\subsection{Bottleneck Analysis}
\label{achievements-ctime}
\begin{figure}[ht]

\centering
\begin{minipage}[b]{0.48\linewidth}
\centering
\includegraphics[width=0.95\textwidth]{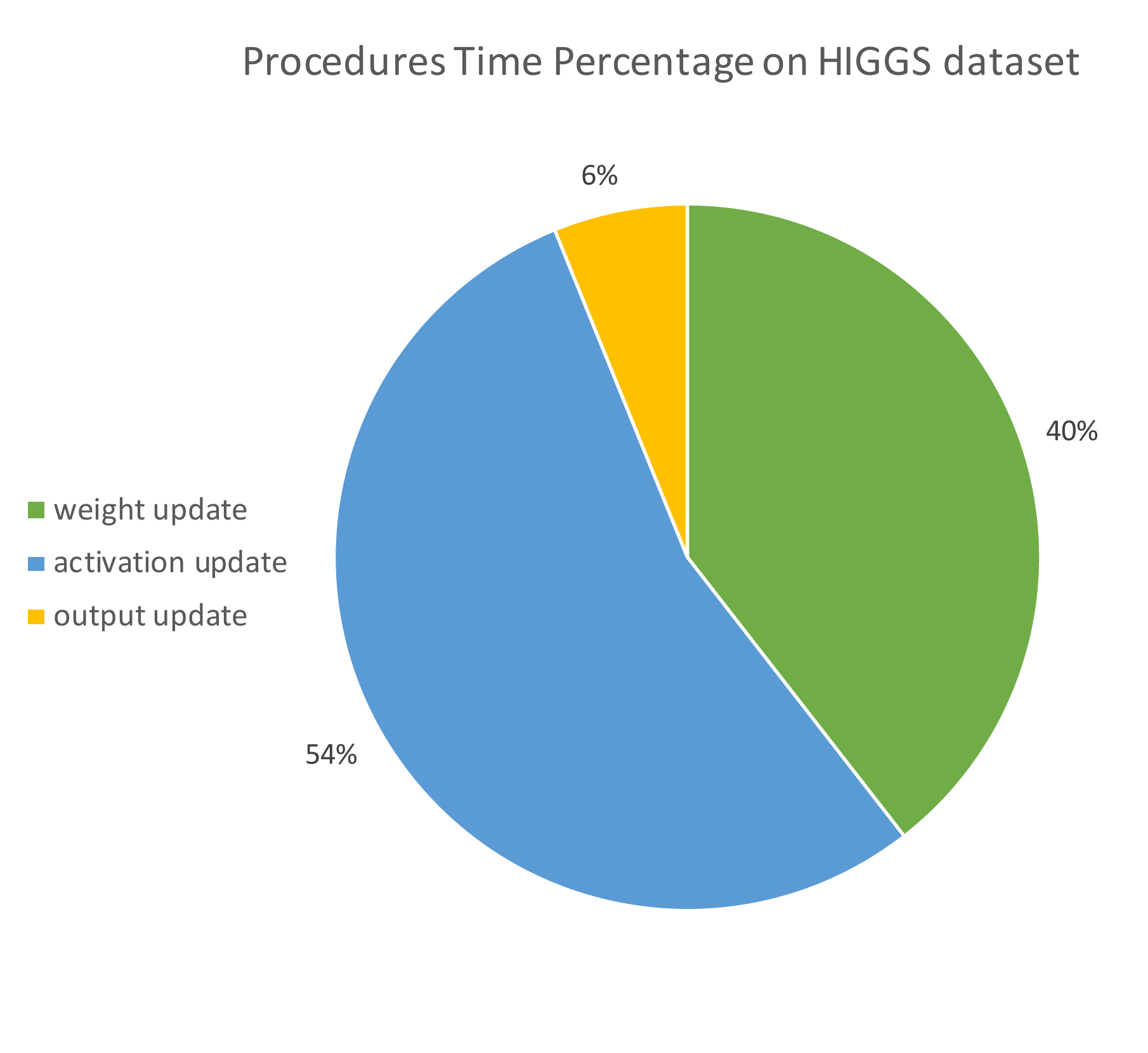}

\end{minipage}
\quad
\begin{minipage}[b]{0.48\linewidth}
\centering

\includegraphics[width=0.95\textwidth]{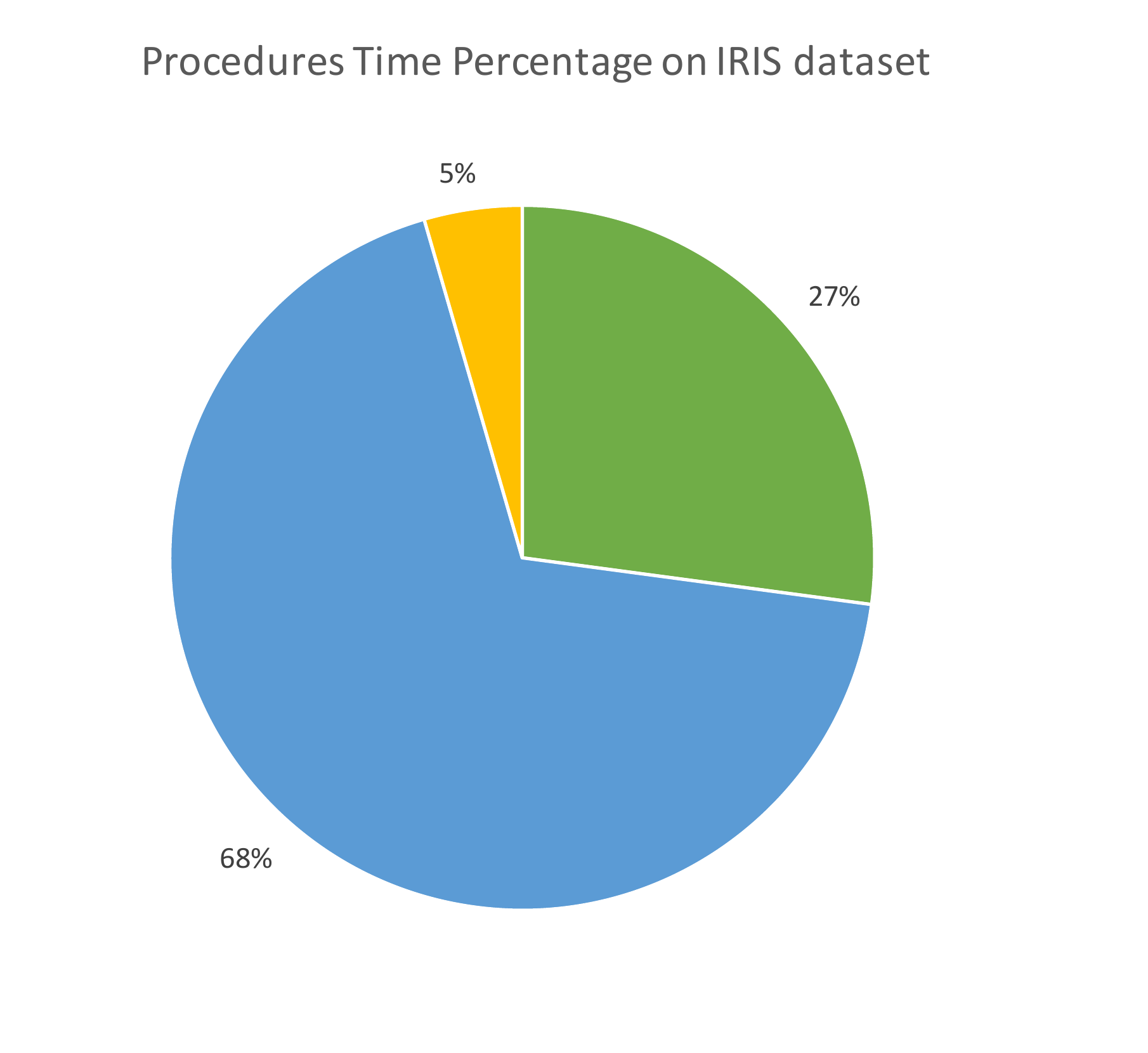}

\end{minipage}
\caption{Piechart of execution time of different procedures in one iteration}
\label{fig:piechart}
\end{figure}

Using the C implementation, we measured the execution time of each procedure in a single iteration of training of a 4 layer neural network with the hidden size of 28 on a subset of HIGGS data set \cite{HIGGS} and a 3 layer neural network with the hidden size of 8 on IRIS data set \cite{IRIS}. These measurements were performed for 3000 iterations. The percentage of execution time associated with each of the procedures is reported in tables \ref{tab:timeHIGGS} and \ref{tab:timeIRIS} and figure \ref{fig:piechart}. As it is evident from the results, the most time-consuming sub-procedures are \textit{\textbf{activation\_update}} and \textit{\textbf{weight\_update}}. Considering algorithm \ref{alg:ADMM-LSMR}, we concluded that these procedures are considerably more time-consuming because of several \textit{\textbf{LSMR}} calls. Therefore, we chose the LSMR function as the primary target for hardware acceleration.

\begin{table}[H]
\def\arraystretch{1.5}%
\caption{Percentage of the execution time of different procedures in one iteration on HIGGS}
\begin{center}
\label{tab:timeHIGGS}
\begin{tabular}{|l|l|l|l|}
\hline
\textit{\textbf{activation\_update}}  & \textit{\textbf{weight\_update}} & \textit{\textbf{output\_update}} & \textit{\textbf{lagrangian update}}  \\ \hline
   54.35 \%           &        39.46 \%         &      6.12 \%        &  0.05 \%                 \\ \hline
\end{tabular}
\end{center}
\end{table}

\begin{table}[H]
\def\arraystretch{1.5}%
\caption{Percentage of execution time of different procedures in one iteration on IRIS}
\begin{center}
\label{tab:timeIRIS}
\begin{tabular}{|l|l|l|l|}
\hline
\textit{\textbf{activation\_update}}  & \textit{\textbf{weight\_update}} & \textit{\textbf{output\_update}} & \textit{\textbf{lagrangian update}} \\ \hline
   68.06 \%        &          29.98 \%          &      4.69\%  &   0.21\%                 \\ \hline
\end{tabular}
\end{center}
\end{table}




 
\newpage
\section{Implementation of Fixed-point Arithmetic}
\label{achievements-fixedpoint}
Fixed-point arithmetic is widely used in FPGA implementation of neural networks, for both inference and training and is considerably faster and more efficient compared to floating-point. In the following sections, we discuss the motivations, challenges and the code structure of our implementation.


\subsection{Motivation}
Since fix-point arithmetic is composed of simpler data types and operations compared to floating-point, it is widely used for general speed up optimisations and especially for hardware designs because it requires less silicon area. Nevertheless, depending on the algorithm, the disadvantage of narrower range and lower precision may adversely affect precision efficiency.\vspace{8pt}\\
In stochastic gradient descent which is a primary version of gradient-based methods, problem space of parameters is explored using small and noisy steps. Such exploration demands relatively high precision during the updates in SGD algorithm. By observing the implemented low-precision algorithms 
\cite{de2017understanding} and also considering the theoretical upper bound on the performance of low-precision SGD \cite{de2015taming} we can conclude that precision-accuracy trade-off has limited the performance of current training algorithms.\vspace{8pt}\\
As the ADMM-LSMR does not involve gradient calculation and by being parallel avoids heavily sequential processes which cause the required precision being accumulated, we conclude that our method would be much less vulnerable to fixed-point errors comparing to SGD. \vspace{8pt}\\
Also it has been proposed that noise in neural networks is a form of regularisation and can help the model to generalise better. This concept has been assessed in previous works like Dropout \cite{srivastava2013improving}  \cite{srivastava2014dropout}, DropConnect \cite{wan2013regularization} and Binary Connect \cite{courbariaux2015binaryconnect}.\vspace{8pt}\\
Considering the above and the fact that fixed-point arithmetic requires less hardware resources and is faster, we deduced that employing this technique will probably result in an overall improvement in hardware-accelerated ADMM-LSMR algorithm. We implemented both 16bit and 32bit fixed-point with four different rounding methods.


\subsection{Challenges}
The challenges in the process of developing the fixed-point arithmetic and embedding it in matrix operations can be summarised as:  
\begin{itemize}[label=$\sqbullet$]
    \item Numerous edge cases to be considered
    \item Keeping track of fraction bits in chain operations
    \item Boundary checks
    \item Temporary containers overflow check and precision
    \item Challenging bit-wise operations and generalisation to support flexible change of fractional bits
    \item Difficult testing procedure
\end{itemize}

\subsection{Fixed-point Arithmetic Details }
First, we defined our fixed-point data type and associated arithmetic functions. Also, we implemented a set of conversion functions for each rounding method.
Next, we implemented a data structure for storing matrices with our fixed-point data type. Finally, we implemented the fixed-point version of all the relevant matrix functions.
In the following sections, details of selected functions are explained. We will only demonstrate the detail for stochastic rounding method as it was the most complicated version. We also only discuss the implementation of 32bit fixed-point. 16bit implementation is similar to the 32bit version with minor differences.

\subsubsection{Fixed-point Data Type}
Arithmetic on fixed-point numbers is almost identical to integers with some adjustment. We used types int32\_t and int16\_t to define 32bit and 16bit fixed-point data types.

\begin{lstlisting}[style=CStyle]
#define WL 32                // Word Length 
#define Fixed int32_t 
#define FL 18                // Fraction Length 
#define IL WL - FL           // Integer Length
#define Epsilon pow(2, - FL) 
#define Ubound_value (float) (pow(2, IL - 1) - pow(2, -FL))
#define Lbound_value (float) (- (1 << (IL - 1)))
#define Ubound (Fixed) ((1LL << (WL - 1)) - 1)
#define Lbound (Fixed) (- (1 << (WL - 1)))
#define ONE_F (Fixed)(1 << FL)
#define MINUS_ONE_F ((Fixed)((1 << FL) - 1)) ^ 0xFFFFFFFF
\end{lstlisting}

We also defined important constants like $WL$ , $FL$ ,$IL$. These constants define the fixed-point data type $fixed\langle WL, FL\rangle$. Another important constant is $\epsilon$. As we mentioned in section \ref{Background-fixedpoint}, $\epsilon$ is the smallest positive number that can be represented given a fixed-point data type.\vspace{8pt}\\
As previously stated in section \ref{Background-fixedpoint}, we should consider both the \textbf{\textit{representation}} and the \textbf{\textit{value}} when working with fixed-point numbers. \textit{Ubound} and \textit{Lbound} define the  \textbf{\textit{representation}} of boundaries in our data type. Since our representation is the same as two's complement, \textit{Ubound} is represented by setting all bits except the leftmost bit to \textit{1}, and the \textit{Lbound} is represented by setting the left most bit to \textit{1} and the others to zero. \textit{Ubound\_value} and \textit{Lbound\_value} store the \textbf{\textit{value}} of boundaries in float. \textbf{\textit{Representation}} of $1$ and $-1$ are also stored in ONE\_F and MINUS\_ONE\_F.\vspace{8pt}\\ 
Considering $fixed\langle 32, 18\rangle$, which is defined above, these constant can be written as below:
\begin{gather}
    \text{Ubound }= 0111 1111 1111 1111 1111 1111 1111 1111 \\
    \text{Lbound }= 1000 0000 0000 0000 0000 0000 0000 0000 \\
    \text{ONE\_F }=  0000 0000 0000 0100 0000 0000 0000 0000 \\
    \text{MINUS\_ONE\_F }= 1111 1111 1111 1100 0000 0000 0000 0000 
\end{gather}
\newpage


To convert a given float number to fixed-point, first, a boundary check is required. This boundary check takes place in \textit{convert} function and either saturation to boundaries is applied, or a specified rounding function is called.

\begin{gather}
    \text{convert(x)} =     
    \begin{cases}
        \text{Ubound    \hspace{10pt} if x $\geq$ Ubound\_value} \\
        \text{Lbound    \hspace{10pt} if x $\leq$ Lbound\_value} \\
        \text{round\_f(x)   \hspace{10pt}  otherwise }   
    \end{cases} 
\end{gather}



Two essential functions were implemented to cast 64bit fixed-point with $FL$ bits of the fraction and $2*FL$ bits of fraction to our defined data type. These two functions are frequently used in the fixed-point arithmetic. The pseudo-code of these functions can be seen in algorithm \ref{alg:cast32}. \vspace{8pt}\\ 
\begin{algorithm}
\SetAlgoLined
\SetKwProg{Fn}{Function}{ }{end}
\Fn{cast\_f64\_simple}
{
\KwIn{int64\_t $x$ } \KwOut{Fixed out}
\If{$ x \leq (\text{int64\_t})$ Lbound }
{
    out $\leftarrow$  Lbound \\
}
\If{$x \geq (\text{int64\_t})$ Ubound }
{
    out $\leftarrow$  Ubound \\
}
\Else
{
    out $\leftarrow$ (Fixed) $x$ \\

}
}
\vspace{5pt}
\Fn{cast\_f64 }
{
\KwIn{int64\_t $x$ }
\KwOut{Fixed out}

    \If{ $x \leq ((\text{int64\_t})$ Lbound$) \ll FL$ }
    {
        out $\leftarrow$ Lbound \\
    }
    \If{ $x \geq ((\text{int64\_t})$ Ubound$) \ll FL$}
    {
        out $\leftarrow$ Ubound\\
    }
    \Else
    {
        diff $\leftarrow (x \& (\text{int64\_t})((1 \ll (FL)) -1 )) $\\
        prob $\leftarrow 1 - \text{diff} * \epsilon $\\
        \If { \text{random} $\leq \text{prob} $}
        {
             out $\leftarrow (\text{Fixed}) (x \gg FL)$\\
        }
        \Else 
        {
             out $\leftarrow ( (\text{Fixed}) (x\gg FL) + 1 )$ \\
        } 
    
    }
}
 \caption{Casting 64bit fixed-point to 32bit fixed-point }
\label{alg:cast32}
\end{algorithm}
The function \textit{cast\_f64\_simple} preforms an overflow check, and then the input is either saturated to boundaries or its leftmost 32bits are discarded. 
 \vspace{8pt}\\
The other function, \textit{cast\_f64}, is more complicated. 
As it can be seen in algorithm \ref{alg:cast32}, first a boundary check is performed. To perform this check, we first cast \textit{Lbound} and \textit{Ubound} to int64\_t, so they become 64 bits and then in order to align them with the input we shift them to left $FL$ times. After the boundary check, if the input is in range, we change it to fit in 32 bit. Since the input number has $2*FL$ bits of fraction, the rightmost $FL$ bits can not be presented in our target data type. We mask these bits using SHIFT and AND operations and store them in diff. We can write:
\begin{gather}
    x - \lfloor x \rfloor = \text{diff} * \epsilon^2
\end{gather}
Where  $\lfloor x \rfloor$ is defined as the largest integer multiple of $\epsilon$ less than or equal to $x$. We use $\text{diff} * \epsilon = \frac{ x - \lfloor x \rfloor}{\epsilon} $ to compute the rounded version of input using \ref{SR} formula.  \vspace{8pt}\\
Three functions were also implemented to perform primary operations on our defined fixed-point data type. The pseudo-code of these functions are provided in algorithm \ref{alg:32fixprimary}.
 \vspace{8pt}\\
\begin{algorithm}
\SetAlgoLined
\SetKwProg{Fn}{Function}{ } {end}
\Fn{add\_f }
{
\KwIn{Fixed $a$, Fixed $b$ }
\KwOut{Fixed out}
    \text{int64\_t }temp $\leftarrow (\text{int64\_t}) a + (\text{int64\_t})b$\\
    cast\_f64\_simple(temp, out)
}
\vspace{5pt}
\Fn{multiply\_f }
{
\KwIn{Fixed $a$, Fixed$ b$ }
\KwOut{Fixed out}
    \text{int64\_t} temp $\leftarrow (\text{int64\_t}) a * (\text{int64\_t}) b$ \\
    cast\_f64(temp, out) \\
}
\vspace{5pt}
\Fn{divide\_f }
{
\KwIn{Fixed $a$, Fixed $b$ }
\KwOut{Fixed out}
    \text{int64\_t }temp $\leftarrow (((\text{int64\_t}) a) \ll FL) / ((\text{int64\_t}) b)$ \\
    cast\_f64(temp, out)\\
}

 \caption{Primary operations on fixed-point data type}
\label{alg:32fixprimary}
\end{algorithm}
For adding two 32 bit fixed-point numbers we simply cast each of them to 64bits then add them to avoid overflow. In the end, \textit{cast\-f\_simple} function is called to fit the result in 32 bits. \vspace{8pt}\\
\newpage
By multiplying two fixed-point numbers, $a$ and $b$, with $FL$ bits of fraction, we have:
\begin{gather}
         \text{ \textbf{\textit{representation}}: } a * b = temp \\ \nonumber
     \text{ \textbf{\textit{value}} : } a * \epsilon * b * \epsilon = temp * \epsilon^2
\end{gather}
So the result has $2*FL$ bits of fractions and we call \textit{cast\_f} function to perform fraction adjustment. \vspace{8pt}\\
In fixed-point by fixed-point division, assuming both numbers having $FL$ bits of fraction, the dividend should be shifted to left $FL$ times. Consider the division $a / b = result $. The value of the result should be:
\begin{gather}
         \text{ result \textbf{\textit{value}}: } \frac{a}{b} 
\end{gather}
Since the \textbf{\textit{value}} of a representation in our data type is calculated as $\textbf{\textit{representation}} * \epsilon $, the representation of the result in our data type should be:
\begin{gather}
      \text{result \textbf{ \textit{representation}}: }  \frac{a}{b * \epsilon} = \frac{a * \epsilon^{-1}}{b} 
\end{gather}

To achieve this we divide $a$ (dividend) by $\epsilon$ using a SHIFT operation.

\subsubsection{Fixed-point Matrix}
In order to use our implemented fixed-point data type in matrix operations, we defined a new struct:
\begin{lstlisting}[style=CStyle]
typedef struct 
{
    int rows;
    int cols;
    Fixed * data;
} fmatrix;
\end{lstlisting}
This struct is the same as the other matrix struct that we defined in section \ref{achievements-c}, except that the type of stored data is \textit{Fixed}. All the associated matrix operations were also implemented for fmatrix. Most of these functions are quite identical to their equivalent version with double data type matrices, with primary operations (add, multiply, divide) being replaced with fixed-point variants.\vspace{8pt}\\
On the other hand, some functions including \textit{matMul}, \textit{dot} and \textit{norm} required more modification. The pseudo-code of these functions are provided in algorithms \ref{alg:32fixmatrixmatmul}, \ref{alg:32fixmatrixdot} and \ref{alg:32fixmatrixnorm} respectively.\vspace{8pt}\\

\begin{algorithm}
\SetAlgoLined
\SetKwProg{Fn}{Function}{ } {end}
\Fn{matMul\_f }
{
\KwIn{$\text{fmatrix } \mat{mat1} \in {\rm I\!R}^{m * n} \text{ , fmatrix } \mat{mat2} \in {\rm I\!R}^{n * p}$}
\KwOut{fmatrix prod}
\For{$ col =1, 2, ..., p $}
{   \For{$ row =1, 2, ..., m$}
    {
        $(\text{int64\_t})  \text{ sum} \leftarrow 0$  \\
        \For{$ k = 1 ,2 ,... , n $ }
        {   
             $(\text{int64\_t})  \text{ temp} \leftarrow (\text{int64\_t}) \mat{mat1}[row][k] * (\text{int64\_t}) \mat{mat2}[k][col]  $\\
             $ \text{ sum} \leftarrow \text{ temp} + \text{ sum} $ \\
            \textit{ overflow check and saturation of} sum \\
             

        }
        
            Fixed result\\
            cast\_f64(sum , result)\\
            prod[row][col] $\leftarrow$ result\\
    }
}

}
 \caption{Matrix multiplication of fmatrix}
\label{alg:32fixmatrixmatmul}
\end{algorithm}
\newpage
In all of these functions MAC (Multiply and Accumulate) operation is performed. To apply this critical operation in fixed-point with minimal error, the following approached was used:
\begin{itemize}[label=$\sqbullet$]
    \item We store the result of each multiplication which is a fixed-point number with $2*FL$ bits of fraction and $2 *IL$ bits of integer in an int64\_t. The conversion of these results to our \textit{Fixed} data type is avoided and is delayed to the next step, but, after each addition, overflow check is applied.  
    \item  We convert the sum of all the results to \textit{Fixed} data type using \textit{cast\_f} function since it is a 64bit fixed-point with $2*FL$ bits of fraction.
\end{itemize}

\begin{algorithm}
\SetAlgoLined
\SetKwProg{Fn}{Function}{ } {end}
\Fn{dot\_f }
{
\KwIn{$\text{fmatrix } \vec{v1} \in {\rm I\!R}^{1 * m} \text{ , fmatrix } \vec{v2} \in {\rm I\!R}^{m * 1}$}
\KwOut{Fixed prod}
 $(\text{int64\_t})  \text{ sum} \leftarrow 0$  \\
\For{$ k =1, 2, ..., m $}
{  
     $(\text{int64\_t})  \text{ temp} \leftarrow (\text{int64\_t}) \mat{v1}[0][k] * (\text{int64\_t}) \mat{v2}[k][0]  $\\
     $ \text{ sum} \leftarrow \text{ temp} + \text{ sum} $ \\
    \textit{ overflow check and saturation of} sum \\
    cast\_f64(sum , prod)\\

}

}
 \caption{Dot product of fmatrix}
\label{alg:32fixmatrixdot}
\end{algorithm}

\begin{algorithm}
\SetAlgoLined
\SetKwProg{Fn}{Function}{ } {end}
\Fn{norm\_f }
{
\KwIn{$\text{fmatrix } \vec{v} \in {\rm I\!R}^{m * 1} $}
\KwOut{Fixed n}
$(\text{int64\_t})  \text{ sum} \leftarrow 0$  \\
 \For{$ row =1, 2, ..., m$}
    {
 
     $(\text{int64\_t})  \text{ temp} \leftarrow (\text{int64\_t}) \vec{v}[row][0] * (\text{int64\_t}) \vec{v}[row][0]  $\\
     $ \text{ sum} \leftarrow \text{ temp} + \text{ sum} $ \\
    \textit{ overflow check and saturation of} sum \\

    }
       
n $\leftarrow$ integer sqrt of sum

}
\caption{L2-norm of fmatrix}
\label{alg:32fixmatrixnorm}
\end{algorithm}
A function to perform integer square root was needed in norm function. We used the proposed algorithm in \cite{intsqrt} with a little modification for this purpose.



\section{Using Fixed-point LSMR in ADMM}
\label{achievement-fixedpointlsmr}
As we discussed in section \ref{achievements-ctime}, LSMR function in \textbf{\textit{weight\_update}} and \textbf{\textit{activation\_update}} was the most time consuming part of the C implementation. In order to speed up the training process, we aimed to make the LSMR function faster by running it on hardware and taking advantage of both task and pipeline parallelism.\vspace{8pt}\\
As previously stated, fixed-point arithmetic is considerably more efficient comparing to floating-point. Therefore, in order to make the hardware implementation more feasible, fixed-point arithmetic was used in the LSMR module. The fixed-point version of LSMR works with fixed-point matrices and uses their relevant functions to perform matrix operations which were explained in section \ref{achievements-fixedpoint}.\vspace{8pt}\\
Also, some experiments were performed to check if the ADMM-LSMR algorithm works with low precision using different rounding methods. Results of these experiments can be found in section \ref{exp-floatvsfixed}.\vspace{8pt}\\
In summary, we observed that:
\begin{itemize}[label=$\sqbullet$]
    \item We were able to achieve near float accuracy using 32bit fixed-point implementation of LSMR.
    \item The proposed ADMM-LSMR method failed to converge using 16bit fixed-point implementation of LSMR. 
\end{itemize}


\chapter{Hardware Design}
\label{ch4}
\section{Hardware-accelerated ADMM-LSMR}
\label{achievements-HWADMMLSMR}
To achieve our final goal, which was hardware implementation of the proposed method, we used Intel FPGA SDK for OpenCL.
A Programmable Acceleration Card with Intel Arria® 10 GX FPGA was used for our design.\vspace{8pt}\\
In this implementation we were able to run \textbf{\textit{weight\_update}} and \textbf{\textit{activation\_update}} procedures in parallel and speed up or training process. To the best of our knowledge, this is the first hardware implementation of ADMM, which also uses LSMR for training neural networks.
\subsection{Motivation}
The proposed ADMM-LSMR method in \cite{iso} is a hardware-friendly approach for training neural networks. From the early stages of this work, our goal was to use hardware acceleration to take advantage of the inherent parallelism of this method. After implementing the fixed-point version of the method and observing its promising results, our next step was to develop an OpenCL program to perform FPGA emulation and finally run on our target FPGA card.\vspace{8pt}\\
In the following sections, the latest version of the implementation, which is a product of extensive optimisation is explained. The details of these optimisation stages are discussed in section \ref{achievements-optimisation}.

\subsection{Challenges}
\begin{itemize}[label=$\sqbullet$]
    \item Extreme system requirements of the development tools. Development had to be performed fully remote on department workstations or Intel Devcloud
    \item Unstable emulator
    \item Lack of informative error/crash reports
    \item Long hardware compilation time.
    \item Getting access to Intel FPGA boards and technical difficulties in working with Intel servers
    \item Minor but undocumented differences between the emulator and physical FPGA
\end{itemize}
\subsection{OpenCL for FPGA Implementation}

As we mentioned in section \ref{Background-opencl}, an OpenCL accelerated program has two parts:
\begin{enumerate}
    \item  A C++ program to run on the host. This program is compiled using g++.
    \item  An OpenCL program including kernels to run on the device which is complied using Intel AOC compiler.
\end{enumerate}
In the following sections, an overview of the program flow and more details of the host and device sections are explained.    

\begin{figure}[H]
\centering
\includegraphics[width = 1\hsize]{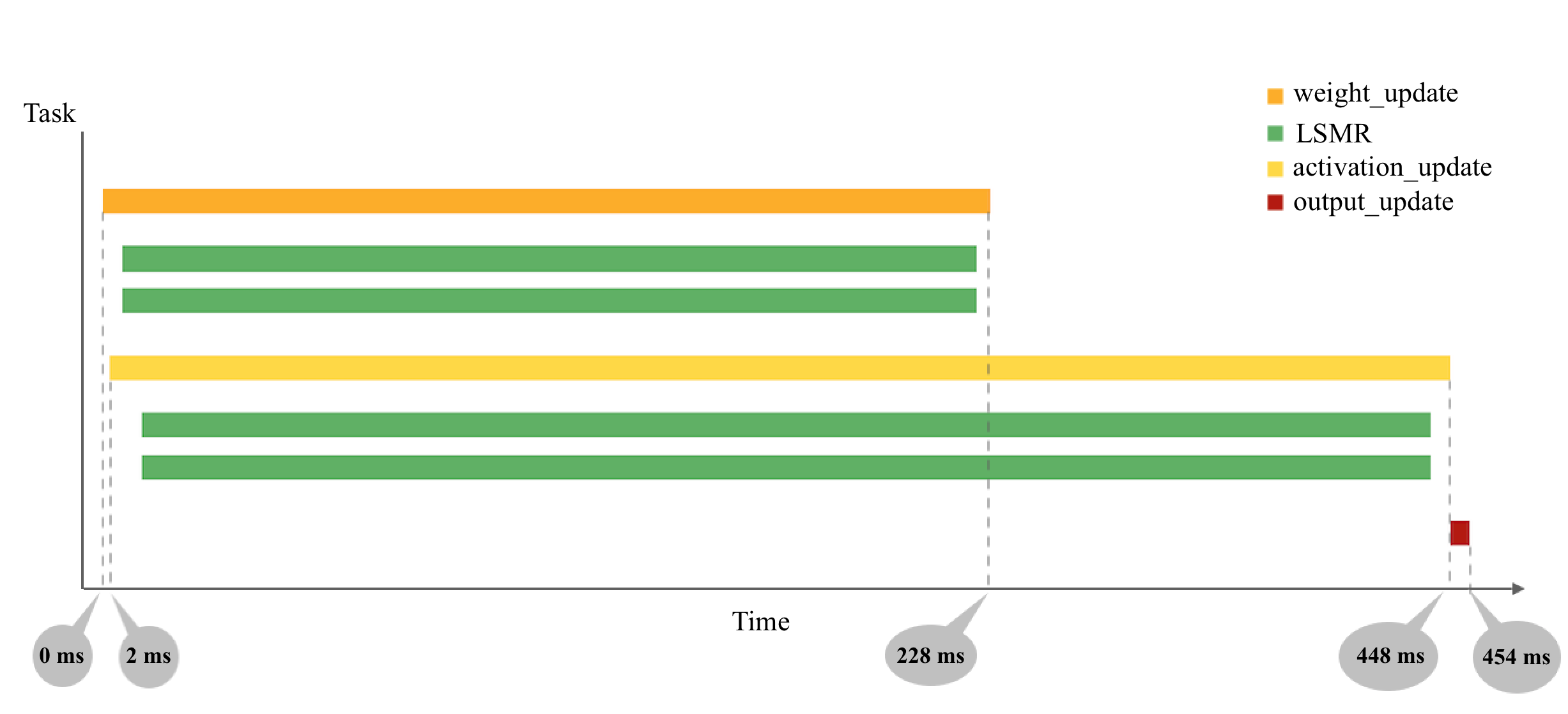}
\caption{Gantt chart of execution time of one iteration of ADMM-LSMR on a hidden layer with hidden size of 28 on HIGGS data set}
\label{fig:gantt}
\end{figure}

\begin{figure}[ht]
\centering
\includegraphics[width = 0.65\hsize]{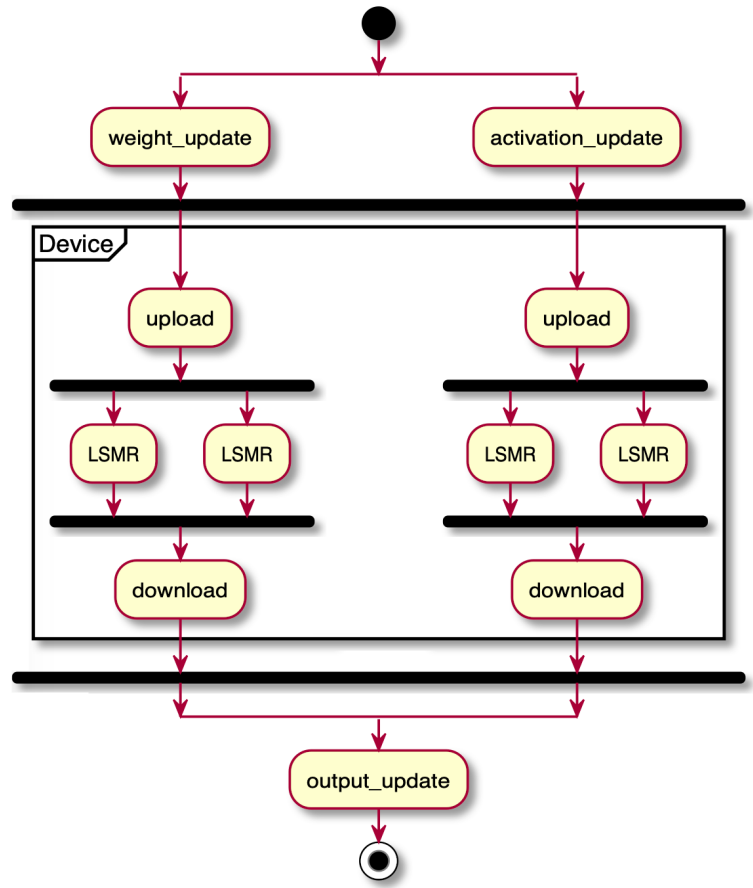}
\caption{Activity diagram of one iteration of ADMM-LSMR on a single layer }
\label{fig:sequence}
\end{figure}
\subsubsection{Program Flow}
A high-level overview of the program can be described as:
\begin{enumerate}
\item Initialising OpenCL run-time and resources
\item Preparing the inputs and setup network architecture
\item Performing ADMM-LSMR until converged as:
\begin{enumerate}
\item Apply the following for each layer:
\begin{enumerate}
\item Performing \textbf{\textit{weight\_update}} and \textbf{\textit{activation\_update}} with LSMR call commands being sent to the FPGA. 
\item Wait for stage \textit{i} results and perform \textbf{\textit{output\_update}} if not in last layer
\end{enumerate}
\item Apply \textbf{\textit{last\_output\_update}} for last layer and \textbf{\textit{lagrangian\_update}}
\end{enumerate}
\item Save the results and clean up the resources
\end{enumerate}

The hardware-accelerated part takes place on stage \textit{3.a}. An activity diagram of this process is shown in figure \ref{fig:sequence}. Also, a Gantt chart illustrating this stage on a single hidden layer of size 28 on HIGGS data set is provided in figure \ref{fig:gantt}.

\subsubsection{Device Program}

As it is shown in pseudo-code \ref{alg:ADMM-LSMR}, in both \textbf{\textit{activation\_update}} and \textbf{\textit{weight\_update}} procedures, the LSMR function is called inside a for loop. This is because pf the fact that original LSMR method solves least-square problems like \ref{linearsystem} and its output has the following form:
\begin{gather}
    x =A^{-1} b  \nonumber \\
    A \in {\rm I\!R}^{m * n}  ,  b \in {\rm I\!R}^{m * 1} ,  x \in {\rm I\!R}^{n * 1} 
\end{gather}
Therefore, in order to solve a problem like \ref{lsmrmatrix}, the LSMR function should be called $p$ times. These $p$ different LSMR calls are independent in relation to each other which makes them an ideal candidate to be implemented in hardware and be parallelised.
\begin{gather}
\label{lsmrmatrix}
    X =A^{-1}B \nonumber \\
    A \in {\rm I\!R}^{m * n}  ,  B \in {\rm I\!R}^{m * p} ,  X \in {\rm I\!R}^{n * p}
\end{gather}

Our implemented LSMR kernel takes 3 matrices as input to solve a problem of a form \ref{lsmrmatrix}. Intel AOC compiler applies pipeline parallelism when it translates the kernel to bitstream for FPGA.
\begin{lstlisting}[style=CStyle]
__kernel void lsmr(const int m, const int n, const int p, __global const Fixed * restrict base_input , __global Fixed * restrict base_At, __global Fixed * restrict base_output, __local Fixed * restrict base_u , __local Fixed * restrict base_v, __local Fixed * restrict base_h ,__local  Fixed * restrict base_hbar, const int offset )
\end{lstlisting}
\vspace{8pt}
As dynamic allocation is not allowed inside the kernel, all of the allocations take place in the host and OpenCL memory objects are passed to the kernel. This memory objects can be seen as a pointer inside the kernel. These memory objects which appear in the header of the kernel are:

\begin{itemize}[label=$\sqbullet$]
    \item \textit{\_\_global const Fixed * restrict base\_input}: Pointer to start of a global memory that stores inputs $A$ and $B$.  We refer to this as input buffer in the host side.
    \item  \textit{\_\_global Fixed * restrict base\_At}: Pointer to starts of a global memory for storing $A^T$ which is a matrix that is needed in internal computing of the kernel. We refer to this as internal buffer in the host side.
    \item  \textit{\_\_global Fixed * restrict base\_output}: Pointer to starts of a global memory for storing the output $X$. We refer to this as output buffer in the host side.
    \item \textit{\_\_local Fixed * restrict base\_u ,\_\_local Fixed * restrict base\_v, \_\_local Fixed * restrict base\_h ,\_\_local  Fixed * restrict base\_hbar}: Pointers to start of local memories for storing internal matrices needed for kernel computations.

\end{itemize}

In addition to these memory objects, the kernel takes four integers.
$m$, $n$ and $p$ determine dimensions of the input and output matrices. \vspace{8pt}\\
In order to utilise most of the available FPGA resources, after the extensive optimisations, we were able to fit \textbf{\textit{four compute units}} of the LSMR kernel in the target FPGA. In every iteration of training, two of these compute units are used for activation update LSMR, and the other two perform weight update LSMR. As we mentioned earlier, computing the columns of the output of LSMR are independent in relation to each other. Therefore in order to split the workload between the compute units, we made each compute unit responsible for computing half of the columns of the output. We used an offset (a const int) to tell the compute units which part of the input they should use and in which part of the output they should write.\vspace{8pt}\\
Another important matter to mention is that this implementation uses round to nearest method for fixed-point numbers. All of the inputs are converted to fixed-point in the host using the specified rounding method in that program. But, the used rounding method inside the kernel is nearest rounding. \vspace{8pt}\\
Since the inputs of LSMR were memory objects, we had to define another version of fmatrix in the kernel program. All of the matrix operation functions had to be modified consequently. Most of these functions are not used explicitly in the final version since we had to make the matrix operations inline to optimise the kernel performance and resource utilisation.
The final implementation is quite complex, tangled and hard to read as a result of hardware optimisations.
\subsubsection{Host Program}
The host program has three main tasks:
\begin{enumerate}
    \item Configure OpenCL runtime and initialising OpenCL objects.
    \item Perform the skeleton of the algorithm.
    \item Orchestrate the acceleration and device commands. 
\end{enumerate}

\subsubsection{\textit{Configuration and Initialisation}}
We defined a manager for OpenCL to perform all of the initial configurations of the OpenCL and also to keep track of important objects including platform, device, context, program and command queues. As we mentioned earlier, we have four compute units that we want to work in parallel. Therefore it was required to define four different command queues.
\begin{lstlisting}[style=CStyle]
typedef struct 
{
  cl_platform_id platform; 
  scoped_array<cl_device_id> device; 
  cl_context context;
  cl_command_queue queue0;
  cl_command_queue queue1;
  cl_command_queue queue2;
  cl_command_queue queue3;
  cl_program program;

} opencl_manager;
\end{lstlisting} \vspace{8pt}
We also defined a function to perform initialisation, \textit{init\_opencl}, and another one to release allocated objects, \textit{cleanup\_opencl}. These two functions are invoked at the start and end of the host program respectively.  \vspace{8pt}\\
In the \textit{init\_opencl} function we perform the following:
\begin{enumerate}
    \item Get the OpenCL platform.
    \item Query the available OpenCL devices and pick the first one.
    \item Create the context.
    \item Create the program and build it.
    \item Create the command queues.
\end{enumerate}

Each LSMR kernel invocation requires a set of parameters and OpenCL objects. Therefore the following struct was defined. This struct includes dimensions of the matrices, all the input, output and other buffers that the kernel requires, and event objects used to define dependencies of commands and synchronisation with the host. \\
\begin{lstlisting}[style=CStyle]
typedef struct 
{
    int m;
    int n;
    int p;
    cl_mem input_buf; 
    cl_mem internal_buf1; 
    cl_mem internal_buf2;
    cl_mem output_buf;
    cl_kernel kernel1;
    cl_kernel kernel2;
    cl_event kernel_event[2];
    cl_event finish_event[2];

} lsmr_module;
\end{lstlisting} 
In the host program we defined two dynamic arrays of \textit{lsmr\_module}. \textit{weight\_update\_lsmr} and \textit{activation\_update\_lsmr}. We constructed one \textit{lsmr\_module} for each call of \textit{weight\_update} and \textit{activation\_update} in one training iteration. As the dimensions of inputs and output of the function in different iterations are the same, these modules are reused in iterations of the training. Therefore \textit{weight\_update\_lsmr} and \textit{activation\_update\_lsmr} arrays are consist of \textit{layers} and \textit{ layers -1} number of \textit{lsmr\_modules} respectively. \vspace{8pt}\\
We also defined a function in order to initialise \textit{lsmr\_modules} with the appropriate parameters. The important parameters are m, n and p which determine the dimensions of the inputs and output of the LSMR.\vspace{8pt}\\
This function performs the following:
\begin{enumerate}
    \item Create input buffer, output buffer and internal buffer with the appropriate size.
    \item Create the kernel and set its arguments.
\end{enumerate}


\subsubsection{\textit{Device Invocation}}

The host program is responsible for managing the device kernel calls. This process includes setting kernel arguments, uploading the inputs, invoking the kernel and downloading the results. It also manages dependencies and synchronisation of upload, invocation, download and in general host-device synchronisation. \vspace{8pt}\\
In our implementation, a function called \textit{run\_lsmr\_opencl} is responsible to perform OpenCL commands given an initialised \textit{lsmr\_module} object. This function gets called from the \textbf{\textit{weight\_update}} and \textbf{\textit{activation\_update}} functions and performs the following actions:
\begin{enumerate}
    \item \textit{clEnqueueWriteBuffer} to upload input matrix $A$.
    \item \textit{clEnqueueWriteBuffer} to upload input matrix $B$, we upload this matrix column wise.
    \item \textit{clEnqueueWriteBuffer} to fill the output buffer with zero.
    \item \textit{clEnqueueTask} to invoke the kernel for computing first half of the result.
    \item \textit{clEnqueueTask} to invoke the kernel for computing second half of the result.
    \item \textit{clEnqueueReadBuffer} to download the output $X$ from the output buffer.
\end{enumerate}
It is worth mentioning that for each \textit{clEnqueueTask}, we had to use a different queue so they would be able to run in parallel. Also, OpenCL events are used to ensure the kernels start after the upload is finished and likewise, the download is started when the kernels are completed.\vspace{8pt}\\
In the main training loop, after calling the \textbf{\textit{weight\_update}} and \textbf{\textit{activation\_update}} a function called post process is invoked where we wait for the results of the two enqueued kernels and perform the further required operations. \vspace{8pt}\\
As previously stated, we use four LSMR compute units in parallel and each of them is pipelined internally. As a result, the two most consuming parts of the training process execute in parallel and pipelined fashion which leads to a noticeable speed up.
\section{Optimisations of FPGA Implementation}
\label{achievements-optimisation}

\
In this section, we describe the applied optimisation steps and their results on timing and resource usage. These optimisations were critical to maximise the performance and utilisation of the available resources. Our goal was to achieve the maximum frequency of our target device (240 MHz), and II equal to one for most parts of the design.\vspace{8pt}\\
In each step, we have provided two tables for showing some of the non-optimised blocks of code and critical issues based on II and fmax values. A separate table is also provided to show the resource usage of each design.
\newpage
\subsubsection{Version 1}
Our first OpenCL implementation of the LSMR kernel was similar to its C implementation with minor modification. This version took as input a matrix $A \in {\rm I\!R}^{m * n}$ and a vector $b \in {\rm I\!R}^{m * 1}$ and produced an output of the form $x \in {\rm I\!R}^{n * 1}$. Some critical deficiencies of this design is shown in tables \ref{tab:v1f} and \ref{tab:v1II}. One of the main issues of this implementation was the memory dependency between the load and store operations which caused a high value of II in different sections of the code. 
\begin{table}[H]
\def\arraystretch{1.25}%
\caption{Non-optimised blocks of design based on II. Version 1. }
\begin{center}
\label{tab:v1II}
\begin{tabular}{|l|l|l|}
\hline
\textbf{Location in source code} & \textbf{II} & \textbf{Details}                                                                                   \\ \hline
\textit{Computing norm}          & $\sim$172   & \begin{tabular}[c]{@{}l@{}}Data dependency. \\ Load from global memory.\end{tabular}               \\ \hline
\textit{Summation of matrices}   & $\sim$258   & \begin{tabular}[c]{@{}l@{}}Memory dependency.\\ Load and then store to global memory.\end{tabular} \\ \hline
\textit{Transposing matrix}      & $\sim$257   & \begin{tabular}[c]{@{}l@{}}Memory dependency.\\ Load and then store to global memory.\end{tabular} \\ \hline
\end{tabular}
\end{center}
\end{table}

\begin{table}[H]
\def\arraystretch{1.25}%
\caption{Non-optimised blocks of design based on fmax. Version 1. }
\begin{center}
\label{tab:v1f}

\begin{tabular}{|l|l|l|}
\hline
\textbf{Location in source code} & \textbf{Scheduled fmax} & \textbf{Details}                                                                                   \\ \hline
\textit{Computing Integer SQRT }          & 98.3    &  Loop feedback             \\ \hline
\textit{ Performing Matrix Multiplication}   & 175.0   & Loop feedback \\ \hline
\textit{Computing norm }      & 	135.0    &  Loop feedback \\ \hline
\end{tabular}

\end{center}
\end{table}
\begin{table}[H]
\def\arraystretch{1.25}%
\caption{Estimated resource of system. Version 1. }
\begin{center}
\label{tab:v1area}

\begin{tabular}{l|l|l|l|l|l|}
\cline{2-6}
& \textbf{ALUTs} & \textbf{FFs}  & \textbf{RAMs} & \textbf{MLABs} & \textbf{DSPs} \\ \hline
\multicolumn{1}{|l|}{\textbf{Board Interface}} & 8\% & 8\% &6\% & 0\%      & 0\%            \\ \hline
\multicolumn{1}{|l|}{\textbf{Kernel System}}  & 28\%                       & 18\%                        & 36\% & 3\%& 33\% \\ \hline
\multicolumn{1}{|l|}{\textbf{Total}}           &36\% & 26\%& 42\% & 3\% & 33\%    \\ \hline
\end{tabular}
\end{center}
\end{table}

\subsubsection{Version 2}
In this version, we changed the LSMR kernel to work on inputs of the form $A \in {\rm I\!R}^{m * n}$ and vector $B \in {\rm I\!R}^{m * p}$ and produce an output of the form $B \in {\rm I\!R}^{n * p}$. We also used local memory for two of internal vectors, which were accessed more frequently than others, to reduce the size of internal buffer and to reduce the number of load and store from the global memory. This modification was done to address one of the issues of the former version. As a general rule, if possible, it is better to copy parts of memory that are accessed more than once into local memory. 
 
\begin{table}[H]
\def\arraystretch{1.25}%
\caption{Non-optimised blocks of design based on II. Version 2. }
\begin{center}
\label{tab:v2II}
\begin{tabular}{|l|l|l|}
\hline
\textbf{Location in source code}                                                                & \textbf{II} & \textbf{Details}                                                                                   \\ \hline
\textit{\begin{tabular}[c]{@{}l@{}}Computing norm of a vector \\ in local memory\end{tabular}}  & $\sim$41    & \begin{tabular}[c]{@{}l@{}}Data dependency. \\ Load from local memory.\end{tabular}                \\ \hline
\textit{\begin{tabular}[c]{@{}l@{}}Computing norm of a vector \\ in global memory\end{tabular}} & $\sim$156   & \begin{tabular}[c]{@{}l@{}}Data dependency. \\ Load from global memory.\end{tabular}               \\ \hline
\textit{\begin{tabular}[c]{@{}l@{}}Summation of two vectors \\ in global memory\end{tabular}}   & $\sim$214   & \begin{tabular}[c]{@{}l@{}}Memory dependency.\\ Load and then store to global memory.\end{tabular} \\ \hline
\end{tabular}

\end{center}
\end{table}



\begin{table}[H]
\def\arraystretch{1.25}%
\caption{Estimated resource usage of system. Version 2. }
\begin{center}
\label{tab:v2area}
\begin{tabular}{l|l|l|l|l|l|}
\cline{2-6}
                                               & \textbf{ALUTs} & \textbf{FFs}  & \textbf{RAMs} & \textbf{MLABs} & \textbf{DSPs} \\ \hline
\multicolumn{1}{|l|}{\textbf{Board Interface}} & 8\%    & 8\%  & 6\%     & 0\%        & 0\%       \\ \hline
\multicolumn{1}{|l|}{\textbf{Kernel System}}   & 32\%  & 20\% & 68\%   & 5\%     & 33\%    \\ \hline
\multicolumn{1}{|l|}{\textbf{Total}}           & 40\%  & 28\% & 74\%   & 5\%     & 33\%    \\ \hline
\end{tabular}
\end{center}
\end{table}

The scheduled fmax value did not change significantly in this step.\vspace{8pt}\\
By comparing tables \ref{tab:v1area} and  \ref{tab:v2area} we can observe that the only major difference in resource usage between these two versions is the RAM usage, even though the version 2 design can perform the same operation as the version 1 but on multiple columns of input using a for loop. 
This comparison demonstrates that the \textbf{\textit{AOC compiler does not replicate the hardware for each column, and tries to apply pipeline parallelism.}} \vspace{8pt}\\
The high amount of RAM usage in version 2 is resolved in the next versions.
\subsubsection{Version 3}
In previous versions, in most of the nested loops, the outer loop was not pipelined. This was because the compiler was not able to recognise that number of iterations of the inner loop are the same for different iterations of the outer loop. In this version, we guided the compiler to pipeline these outer loops by using constant type copies of the variables for specifying the number of loop iterations.\vspace{8pt}\\
Also in previous versions, there was a compile warning about not using \textit{"restrict"} keyword for pointers to memories in kernel signature. This keyword was used in this version which helps the compiler with some cache optimisations and also restricts the effect of pointer aliasing.\vspace{8pt}\\
Another technique used in this step was coalescing nested loops manually wherever possible. In coalescing a nested loop, we transform it into a single loop without changing its functionality. This technique reduces the loop overhead and also latency, and as a result, reduces the kernel resource usage.\vspace{8pt}\\
Fusing adjacent loops is also applied manually in this step. This technique also reduces the loop overhead and therefore the area usage. The main effect of this technique is running the adjacent loops concurrently as they are considered a single loop and this increases the performance. In this version, we managed to use one nested loop instead of three, for calculating \ref{av} as well as calculating \ref{au} which correspond to lines 10 and 12 of algorithm \ref{alg:LSMR}. This was achieved by fusing the following loops: matrix-vector multiplication nested loop, vector-scalar multiplication loop and subtracting vectors loop for both calculations. This also helped with precision and reduced the amount of internal buffer needed by eliminating some internal vectors.

\begin{gather}
\label{av}
    \vec{u_{k+1}} \leftarrow (\mat{A} \vec{v_k} - \alpha_k \vec{u_k} )
\\
\label{au}
 \vec{v_{k+1}} \leftarrow (\mat{A^T} \vec{u_{k+1}} - \beta_{k+1} \vec{v_k} )
\end{gather}
Again the scheduled fmax did not change significantly in this step.\vspace{8pt}\\
The II value changed a little for computing norm of a vector in global memory and summation of vectors.\vspace{8pt}\\
There was a considerable change in the amount of resource usage specially RAM usage. This reduction in resource usage is because of eliminating the overhead of some loops.
\begin{table}[H]
\def\arraystretch{1.25}%
\caption{Non-optimised blocks of design based on II. Version 3. }
\begin{center}
\label{tab:v3II}
\begin{tabular}{|l|l|l|}
\hline
\textbf{Location in source code}                                                                & \textbf{II} & \textbf{Details}                                                                                   \\ \hline
\textit{\begin{tabular}[c]{@{}l@{}}Computing norm of a vector \\ in local memory\end{tabular}}  & $\sim$41    & \begin{tabular}[c]{@{}l@{}}Data dependency. \\ Load from local memory.\end{tabular}                \\ \hline
\textit{\begin{tabular}[c]{@{}l@{}}Computing norm of a vector \\ in global memory\end{tabular}} & $\sim$150   & \begin{tabular}[c]{@{}l@{}}Data dependency. \\ Load from global memory.\end{tabular}               \\ \hline
\textit{\begin{tabular}[c]{@{}l@{}}Summation of two vectors \\ in global memory\end{tabular}}   & $\sim$196   & \begin{tabular}[c]{@{}l@{}}Memory dependency.\\ Load and then store to global memory.\end{tabular} \\ \hline
\end{tabular}
\end{center}
\end{table}

\begin{table}[H]
\def\arraystretch{1.25}%
\caption{Estimated resource usage of system. Version 3. }
\begin{center}
\label{tab:v3area}
\begin{tabular}{l|l|l|l|l|l|}
\cline{2-6}
                                               & \textbf{ALUTs} & \textbf{FFs} & \textbf{RAMs} & \textbf{MLABs} & \textbf{DSPs} \\ \hline
\multicolumn{1}{|l|}{\textbf{Board Interface}} & 8\%            & 8\%          & 6\%           & 0\%            & 0\%           \\ \hline
\multicolumn{1}{|l|}{\textbf{Kernel System}}   & 27\%           & 17\%         & 50\%          & 5\%            & 28\%          \\ \hline
\multicolumn{1}{|l|}{\textbf{Total}}           & 35 \%            & 25\%         & 56 \%           & 5\%            & 28\%          \\ \hline
\end{tabular}
\end{center}
\end{table}

\subsubsection{Version 4}
In this step, more loop fusing was performed. We fused the loop for computing the norm of vector to its adjacent loop, which itself was the result of a loop fusion in the previous step. By this modification, we were able to perform \ref{avn} in one nested loop and \ref{aun} in another nested loop. \ref{avn} and \ref{aun} correspond to lines 10 and 12 of algorithm \ref{alg:LSMR}.

\begin{gather}
\label{avn}
    \vec{u_{k+1}} \leftarrow (\mat{A} \vec{v_k} - \alpha_k \vec{u_k} )
    \\
    \beta_{k+1} \leftarrow ||\mat{A} \vec{v_k} - \alpha_k \vec{u_k}||_2
\end{gather}
\begin{gather}
\label{aun}
 \vec{v_{k+1}} \leftarrow (\mat{A^T} \vec{u_{k+1}} - \beta_{k+1} \vec{v_k} )  \\
 \alpha_{k+1} \leftarrow ||\mat{A^T} \vec{u_{k+1}} - \beta_{k+1} \vec{v_k}||_2
\end{gather}
Also we fused the loops for computing \ref{hbark}, \ref{xk} and \ref{hk} (corresponding to lines 17, 18 and 19 of the algorithm \ref{alg:LSMR}) and we used just one loop to perform all of them.
\begin{gather}
    \label{hbark}
     \vec{\overline{h_{k+1}}} \leftarrow - (\overline{s_k} * \rho_{k+1} * \rho_{k+1}) / ( \rho_k * \overline{\rho_k} ) \vec{\overline{h_k}} + \vec{h_k}  \\
        \label{xk}
     \vec{x_{k+1}} \leftarrow (\zeta_{k+1} / (\rho_{k+1} * \overline{\rho_{k+1}}) \vec{\overline{h_{k+1}}} + \vec{x_k}  \\\label{hk}
      \vec{h_{k+1}} \leftarrow - ((s_{k+1} * \alpha_{k+1}) / \rho_{k+1}) \vec{h_k} + \vec{v}
\end{gather}
These loop fusions again helped with reducing resource usage.
The only factor that changed considerably in this step of optimisation was resource usage:
\begin{table}[H]
\def\arraystretch{1.25}%
\caption{Estimated resource usage of system. Version 4. }
\begin{center}
\label{tab:v4area}
\begin{tabular}{l|l|l|l|l|l|}
\cline{2-6}
                                               & \textbf{ALUTs} & \textbf{FFs} & \textbf{RAMs} & \textbf{MLABs} & \textbf{DSPs} \\ \hline
\multicolumn{1}{|l|}{\textbf{Board Interface}} & 8\%            & 8\%          & 6\%           & 0\%            & 0\%           \\ \hline
\multicolumn{1}{|l|}{\textbf{Kernel System}}   & 25\%           & 15\%         & 37\%          & 4\%            & 28\%          \\ \hline
\multicolumn{1}{|l|}{\textbf{Total}}           & 33\%           & 22\%         & 44\%          & 4\%            & 28\%          \\ \hline
\end{tabular}
\end{center}
\end{table}
\subsubsection{Version 5}
In this version, we used one of the OpenCL built-in integer functions, \textit{add\_sat}. Using this function, eliminated a great number of conditional statements inside the main loop. which were related to overflow check logic. This change, led to a significant reduction in resource usage. \vspace{8pt}\\
We also reduced the value of II significantly and as it can be seen in table \ref{tab:v5II}, the only block of code with value of II bigger than 1 was the square root block. The value of II for integer square root was 4 in all the other versions but not mentioned in the tables as there were more critical blocks in previous versions. Again, the scheduled fmax did not change significantly.

\begin{table}[H]
\def\arraystretch{1.25}%
\caption{Non-optimised blocks of design based on II. Version 5. }
\begin{center}
\label{tab:v5II}
\begin{tabular}{|l|l|l|}
\hline
\textbf{Location in source code} & \textbf{II} & \textbf{Details} \\ \hline
\textit{Integer SQRT}            & 4           & Data dependency. \\ \hline
\end{tabular}
\end{center}
\end{table}

\begin{table}[H]
\def\arraystretch{1.25}%
\caption{Estimated resource of system. Version 5. }
\begin{center}
\label{tab:v5area}
\begin{tabular}{l|l|l|l|l|l|}
\cline{2-6}
                                               & \textbf{ALUTs} & \textbf{FFs} & \textbf{RAMs} & \textbf{MLABs} & \textbf{DSPs} \\ \hline
\multicolumn{1}{|l|}{\textbf{Board Interface}} & 8\%            & 8\%          & 6\%           & 0\%            & 0\%           \\ \hline
\multicolumn{1}{|l|}{\textbf{Kernel System}}   & 11\%           & 7\%          & 20\%          & 3\%            & 18\%          \\ \hline
\multicolumn{1}{|l|}{\textbf{Total}}           & 19\%           & 15\%         & 26\%          & 3\%            & 18\%          \\ \hline
\end{tabular}
\end{center}
\end{table}

\subsubsection{Version 6}
To solve to problem of II value of integer square root block, we used the OpenCL built-in function for computing the float square root. In order to use this function, we had to convert the input from fixed-point to float and convert the output back to fixed-point. By this change, we were able to reduce the II value to 1 as the Intel implementation of the built-in square root function is highly optimised. However, we sacrificed a small amount of RAM usage (Less than one percent) that is negligible. This also solved the problem of low fmax in integer sqrt and we were able to increase the fmax value from 98 (as it is shown in table \ref{tab:v1f}) to 240 for this block \vspace{8pt}\\
In this version all of the blocks of code shown the II value of (1 , \~ 1 , $\geq$ 1) based on the report. 
We also used \textit{prefetch\_load} in this version for reading from the global memory.
\begin{table}[H]
\def\arraystretch{1.25}%
\caption{Estimated resource of system. Version 6. }
\begin{center}
\label{tab:v6area}
\begin{tabular}{l|l|l|l|l|l|}
\cline{2-6}
                                               & \textbf{ALUTs} & \textbf{FFs} & \textbf{RAMs} & \textbf{MLABs} & \textbf{DSPs} \\ \hline
\multicolumn{1}{|l|}{\textbf{Board Interface}} & 8\%            & 8\%          & 6\%           & 0\%            & 0\%           \\ \hline
\multicolumn{1}{|l|}{\textbf{Kernel System}}   & 11\%           & 7\%          & 20\%          & 3\%            & 18\%          \\ \hline
\multicolumn{1}{|l|}{\textbf{Total}}           & 19\%           & 15\%         & 26\%          & 3\%            & 18\%          \\ \hline
\end{tabular}
\end{center}
\end{table}
\subsubsection{Version 7}
In this version we replaced the \textit{add\_sat} with add and we observed that this alteration did not lead to any loss in accuracy. Instead, we were able to solve the problem of low fmax values. In this version, all of the code blocks were showing the scheduled fmax equal to \textit{240 MHz}, which is the maximum achievable frequency on our target board. The area usage did not change significantly.\vspace{8pt}\\
In this version, optimisation was completed. In the next versions, we focused on achieving maximum hardware utilisation. 
\subsubsection{Version 8}
At this stage, we used two LSMR compute units to run the LSMR in the \textbf{\textit{activation\_update}} and \textbf{\textit{weight\_update}} in parallel. It can be observed in table \ref{tab:v8area} that there were still unused hardware resources.
\begin{table}[H]
\def\arraystretch{1.25}%
\caption{Estimated resource of system. Version 8. }
\begin{center}
\label{tab:v8area}
\begin{tabular}{l|l|l|l|l|l|}
\cline{2-6}
                                               & \textbf{ALUTs} & \textbf{FFs} & \textbf{RAMs} & \textbf{MLABs} & \textbf{DSPs} \\ \hline
\multicolumn{1}{|l|}{\textbf{Board Interface}} & 8\%            & 8\%          & 6\%           & 0\%            & 0\%           \\ \hline
\multicolumn{1}{|l|}{\textbf{Kernel System}}   & 23\%           & 14\%         & 40\%          & 6\%            & 37\%          \\ \hline
\multicolumn{1}{|l|}{\textbf{Total}}           & 31\%           & 22\%         & 46\%          & 6\%            & 37\%          \\ \hline
\end{tabular}
\end{center}
\end{table}
\subsubsection{Version 9}
To take advantage of the remaining available hardware resources, we increased the number of compute units of LSMR kernel to 4 on our FPGA. Consequently, the host code was modified to split the workload of LSMR in each of the \textbf{\textit{activation\_update}} and \textbf{\textit{weight\_update}} between two compute units. Also, the LSMR kernel was modified to only work on a segment of the problem to fit into this purpose. \vspace{8pt}\\
This modification led to about 2X speed up compared to the previous version as we doubled the task parallelism.
\vspace{8pt}\\
It is also worth mentioning that the board interface resource usage, which was the same in all of the steps, slightly increased in this step.

\begin{table}[H]
\def\arraystretch{1.25}%
\caption{Estimated resource of system. Version 9. }
\begin{center}
\label{tab:v9area}
\begin{tabular}{l|l|l|l|l|l|}
\cline{2-6}
                                               & \textbf{ALUTs} & \textbf{FFs} & \textbf{RAMs} & \textbf{MLABs} & \textbf{DSPs} \\ \hline
\multicolumn{1}{|l|}{\textbf{Board Interface}} & 16\%           & 10\%         & 15\%          & 0\%            & 0\%           \\ \hline
\multicolumn{1}{|l|}{\textbf{Kernel System}}   & 46\%           & 27\%         & 81\%          & 11\%           & 75\%          \\ \hline
\multicolumn{1}{|l|}{\textbf{Total}}           & 62\%           & 37\%         & 96\%          & 11\%           & 75\%          \\ \hline
\end{tabular}
\end{center}
\end{table}
\chapter{Experimental Results}
\label{ch5}
In this project, two data sets were used for experiments. IRIS \cite{IRIS} which is a small data set and a subset of HIGGS which is a bigger and more complex data set. The achieved test accuracy of the ADMM-LSMR method was assessed against two state-of-the-art gradient-based methods in \cite{iso}. It was observed that the ADMM-LSMR algorithm is able to achieve better accuracy compared to SGD and Adam on HIGGS and IRIS data sets. In this work, first, we compared the achieved test accuracy of the implemented method when using fixed-point arithmetic in LSMR against the original floating-point implementation. Second, we compared the execution time of each training iteration of C implementation and hardware-accelerated FPGA implementation. Finally, we studied the impact of increasing hidden size on the execution time of each iteration in both CPU and FPGA accelerated implementations. \vspace{8pt}\\
The key observation can be summarised as the following:
\begin{itemize}[label=$\sqbullet$]
    \item We were able to achieve near floating point accuracy, with less than one percent penalty using fixed point LSMR with nearest rounding method.
    \item The nearest rounding method was the best choice when using fixed-point version of LSMR in ADMM. This was an important observation as it is reported that using the stochastic rounding in gradient-based methods has the best accuracy compared to other rounding methods \cite{gupta2015deep}.
    \item We were able to achieve up to 6X speed up when using the hardware-accelerated FPGA implementation compared to the C implementation.
    \item The speed up gain of using hardware-accelerated implementation grows by increasing the hidden size of the network.
\end{itemize}

The experiments were conducted on two different machines. For CPU runs, one of the custom computing lab workstations (cccad5) was used and for FPGA accelerated runs Intel Devcloud nodes were employed. The specification of these platforms were the following: 
\begin{description}
\item \textbf{cccad5 workstation}\hfill

CPU model name: Intel(R) Xeon(R) Gold 6154 CPU @ 3.00GHz\\
CPU cache size: 25344 KB\\
CPU cores: 18\\
Memory: 768GB

\item \textbf{Intel Devcloud node}\hfill

CPU model name: Intel(R) Xeon(R) Gold 6230 CPU @ 2.10GHz\\
CPU cache size: 28160 KB\\
CPU cores: 20\\
Memory: 18 GB\\ 
FPGA Board : Intel® Programmable Acceleration Card with Intel Arria® 10 GX FPGA\\
\end{description}

\section{Comparing Floating-point vs Fixed-point}
\label{exp-floatvsfixed}
In this section, we compared the accuracy of training algorithm using fixed-point and floating-point arithmetic. The results were derived from 500 runs of each training algorithm.
\subsection{IRIS}
\subsubsection{3 layer network with hidden size equal to 8}
In this section, a 3 layer network with hidden size equal to 8 is used to train on IRIS data set using the ADMM-LSMR algorithm. We trained this model using fixed-point LSMR with four different rounding methods and floating-point LSMR. \\
\begin{table}[ht]
\def\arraystretch{1.25}%
\caption{Comparing accuracy of using floating-point vs fixed-point with different rounding methods on IRIS }
\begin{center}
\label{tab:IRISfixedvsfloat}
\begin{tabular}{l|l|l|}
\cline{2-3}
\textbf{}                                                           & \textbf{Mean} & \textbf{STDV} \\ \hline
\multicolumn{1}{|l|}{\textit{Floating-point}}                       &       83.7 \%  & 1.6\%             \\ \hline
\multicolumn{1}{|l|}{\textit{Fixed-point with nearest rounding}}    &       82.8 \%& 2.0\%           \\ \hline
\multicolumn{1}{|l|}{\textit{Fixed-point with stochastic rounding}} &       82.3 \% &    3.1\%           \\ \hline
\multicolumn{1}{|l|}{\textit{Fixed-point with upward rounding}}     &         81.1\%       &  5.3\%             \\ \hline
\multicolumn{1}{|l|}{\textit{Fixed-point with downward rounding}}   &       80.8\%         &   5.2\%            \\ \hline
\end{tabular}
\end{center}
\end{table}



We observed that the average of achieved accuracy using nearest rounding is better than the other rounding methods. Also, the achieved accuracy of this method has less variance compared to other rounding methods.\vspace{8pt}\\
It is worth mentioning that as opposed to the conventional training methods such as SGD which perform better using the stochastic rounding method with fixed-point, we did not observe significant difference by using this rounding method.\vspace{8pt}\\
We also observed less than 1\% penalty in accuracy when using fixed-point with nearest rounding method compared to floating-point.

\subsection{HIGGS}
In this section, we trained three neural networks with hidden sizes of 8, 14 and 28 on a subset of HIGGS data set using ADMM-LSMR algorithm. We trained these models using fixed-point LSMR with four different rounding methods and also using floating-point LSMR. 
\subsubsection{3 layer network with hidden size equal to 8}

\begin{table}[H]
\def\arraystretch{1.25}%
\caption{Comparing accuracy of using floating-point vs fixed-point with different rounding methods on HIGGS }
\begin{center}
\label{tab:HIGGS188fixedvsfloat}
\begin{tabular}{l|l|l|}
\cline{2-3}
\textbf{}                                                           & \textbf{Mean} & \textbf{STDV} \\ \hline
\multicolumn{1}{|l|}{\textit{Floating-point}}                       &       63.6 \%        &     0.1\%           \\ \hline
\multicolumn{1}{|l|}{\textit{Fixed-point with nearest rounding}}    &       62.8 \%         &    0.2\%           \\ \hline
\multicolumn{1}{|l|}{\textit{Fixed-point with stochastic rounding}} &       62.6 \%       &           0.2\%     \\ \hline
\multicolumn{1}{|l|}{\textit{Fixed-point with upward rounding}}     &         62.6\%       &    0.1\%            \\ \hline
\multicolumn{1}{|l|}{\textit{Fixed-point with downward rounding}}   &       62.6\%         &      0.2\%          \\ \hline
\end{tabular}
\end{center}
\end{table}



\subsubsection{3 layer network with hidden size equal to 14}

\begin{table}[H]
\def\arraystretch{1.25}%
\caption{Comparing accuracy of using floating-point vs fixed-point with different rounding methods on HIGGS }
\begin{center}
\label{tab:HIGGS11414fixedvsfloat}
\begin{tabular}{l|l|l|}
\cline{2-3}
\textbf{}                                                           & \textbf{Mean} & \textbf{STDV} \\ \hline
\multicolumn{1}{|l|}{\textit{Floating-point}}                       &       63.6 \%        &     0.1\%           \\ \hline
\multicolumn{1}{|l|}{\textit{Fixed-point with nearest rounding}}    &       62.7 \%         &    0.2 \%           \\ \hline
\multicolumn{1}{|l|}{\textit{Fixed-point with stochastic rounding}} &       62.7 \%       &           0.1\%     \\ \hline
\multicolumn{1}{|l|}{\textit{Fixed-point with upward rounding}}     &         62.6\%       &    0.1\%            \\ \hline
\multicolumn{1}{|l|}{\textit{Fixed-point with downward rounding}}   &       62.7\%         &      0.2\%          \\ \hline
\end{tabular}
\end{center}
\end{table}



\subsubsection{3 layer network with hidden size equal to 28}

\begin{table}[H]
\def\arraystretch{1.25}%
\caption{Comparing accuracy of using floating-point vs fixed-point with different rounding methods on HIGGS }
\begin{center}
\label{tab:HIGGS12828fixedvsfloat}
\begin{tabular}{l|l|l|}
\cline{2-3}
\textbf{}                                                           & \textbf{Mean} & \textbf{STDV} \\ \hline
\multicolumn{1}{|l|}{\textit{Floating-point}}                       &       61.3 \%        &     0.1\%           \\ \hline
\multicolumn{1}{|l|}{\textit{Fixed-point with nearest rounding}}    &       62.8 \%         &    0.2 \%           \\ \hline
\multicolumn{1}{|l|}{\textit{Fixed-point with stochastic rounding}} &       62.8 \%       &           0.1\%     \\ \hline
\multicolumn{1}{|l|}{\textit{Fixed-point with upward rounding}}     &         62.7\%       &    0.2\%            \\ \hline
\multicolumn{1}{|l|}{\textit{Fixed-point with downward rounding}}   &       62.7\%         &      0.2\%          \\ \hline
\end{tabular}
\end{center}
\end{table}

We observed that the average of achieved accuracy using nearest rounding is better than the other rounding methods.\vspace{8pt}\\
It is worth mentioning that as opposed to the conventional training methods such as SGD which perform better using the stochastic rounding method with fixed-point, we did not observe significant difference by using this rounding method.\vspace{8pt}\\
We also observed less than 1\% penalty in accuracy when using fixed-point with nearest rounding method compared to floating-point.\vspace{8pt}\\
Additionally, it is evident that by increasing the hidden size of the network, the floating-point is subject to minor loss in accuracy while such behaviour is not observed in fixed-point LSMR implementation. This can be due to the fact that using fixed-point injects noise to the neural network, which delays the overfitting and helps the network to generalise better. 
\section{Comparing CPU Implementation and FPGA Implementation: Accuracy}
\label{exp-cpuvsfpga-accuracy}
In this section, we compared the accuracy of FPGA implementation with CPU implementation. The results are from running each training algorithm 500 times.\vspace{8pt}\\
As expected, the FPGA implementation was able to achieve the same accuracy as the C implementation. This set of experiments were done to assess the correctness of the FPGA implementation. 
\subsection{IRIS}
\subsubsection{3 layer network with hidden size equal to 8}
\begin{table}[H]
\def\arraystretch{1.25}%
\caption{Comparing accuracy of C implementation vs FPGA implementation }
\begin{center}
\label{tab:IRISCPUvsFPGA}
\begin{tabular}{l|l|l|}
\cline{2-3}

\textbf{}                                                           & \textbf{Mean} & \textbf{STDV} \\ \hline
\multicolumn{1}{|l|}{\textit{CPU implementation}}                       &       82.8 \%  & 2.0\%             \\ \hline
\multicolumn{1}{|l|}{\textit{FPGA implementation}}    &       82.7 \%& 2.4\%           \\ \hline

\end{tabular}
\end{center}
\end{table}


\subsection{HIGGS}
\subsubsection{3 layer network with hidden size equal to 8}

\begin{table}[H]
\def\arraystretch{1.25}%
\caption{Comparing accuracy of C implementation vs FPGA implementation }
\begin{center}
\label{tab:HIGGSCPUvsFPGA188}
\begin{tabular}{l|l|l|}
\cline{2-3}

\textbf{}                                                           & \textbf{Mean} & \textbf{STDV} \\ \hline
\multicolumn{1}{|l|}{\textit{CPU implementation}}                       &       62.8 \%  & 0.2\%             \\ \hline
\multicolumn{1}{|l|}{\textit{FPGA implementation}}    &       62.6 \%& 0.2\%           \\ \hline

\end{tabular}
\end{center}
\end{table}


\subsubsection{3 layer network with hidden size equal to 14}

\begin{table}[H]
\def\arraystretch{1.25}%
\caption{Comparing accuracy of C implementation vs FPGA implementation }
\begin{center}
\label{tab:HIGGSCPUvsFPGA11414}
\begin{tabular}{l|l|l|}
\cline{2-3}

\textbf{}                                                           & \textbf{Mean} & \textbf{STDV} \\ \hline
\multicolumn{1}{|l|}{\textit{CPU implementation}}                       &       62.7 \%  & 0.2\%             \\ \hline
\multicolumn{1}{|l|}{\textit{FPGA implementation}}    &       62.7 \%& 0.1\%           \\ \hline

\end{tabular}
\end{center}
\end{table}


\subsubsection{3 layer network with hidden size equal to 28}

\begin{table}[H]
\def\arraystretch{1.25}%
\caption{Comparing accuracy of C implementation vs FPGA implementation }
\begin{center}
\label{tab:HIGGSCPUvsFPGA12828}
\begin{tabular}{l|l|l|}
\cline{2-3}
\textbf{}                                                           & \textbf{Mean} & \textbf{STDV} \\ \hline
\multicolumn{1}{|l|}{\textit{CPU implementation}}                       &       62.8 \%  & 0.2\%             \\ \hline
\multicolumn{1}{|l|}{\textit{FPGA implementation}}    &       62.8 \%& 0.2\%           \\ \hline

\end{tabular}
\end{center}
\end{table}
\section{Comparing CPU Implementation and FPGA Implementation: Time}
\label{exp-cpuvsfpga-time}
In this section, we compare the execution time of each loop iteration of implemented ADMM-LSMR algorithm in CPU and FPGA. The execution time of 2500 iterations of each algorithm has been measured to produce these results. A subset of HIGGS data set was used for these experiments.\vspace{8pt}\\
As it is evident, we were able to achieve up to 6 times speed up depending on the architecture of the network.
\subsection{HIGGS}
\subsubsection{3 layer network with hidden size equal to 8}

\begin{table}[H]
\def\arraystretch{1.25}%
\caption{Comparing execution time of C implementation vs FPGA implementation }
\begin{center}
\label{tab:HIGGSCPUvsFPGATIME188}



\begin{tabular}{l|l|l|c|}
\cline{2-4}
\textbf{}                                         
& \textbf{Mean} & \textbf{STDV} & \textbf{Speed up} \\ \hline
\multicolumn{1}{|l|}{\textit{CPU Implementation}}  
&         589.8 $ms$ & 13.3 $ms$        &   \multirow{2}{*}{\textbf{4.1}} \\ \cline{1-3}
\multicolumn{1}{|l|}{\textit{FPGA Implementation}}
&               143.7 $ms$ & 0.4 $ms$              &                   \\ \hline
\end{tabular}

\end{center}
\end{table}

\subsubsection{3 layer network with hidden size equal to 14}

\begin{table}[H]
\def\arraystretch{1.25}%
\caption{Comparing execution time of C implementation vs FPGA implementation }
\begin{center}
\label{tab:HIGGSCPUvsFPGATIME11414}



\begin{tabular}{l|l|l|c|}
\cline{2-4}
\textbf{}                                         
& \textbf{Mean} & \textbf{STDV} & \textbf{Speed up} \\ \hline
\multicolumn{1}{|l|}{\textit{CPU Implementation}}  
&          1391.4 $ms$ & 24.0 $ms$       &   \multirow{2}{*}{\textbf{5}} \\ \cline{1-3}
\multicolumn{1}{|l|}{\textit{FPGA Implementation}}
&               277.7 $ms$ & 0.9 $ms$            &                   \\ \hline
\end{tabular}

\end{center}
\end{table}

\subsubsection{3 layer network with hidden size equal to 28}

\begin{table}[H]
\def\arraystretch{1.25}%
\caption{Comparing execution time of C implementation vs FPGA implementation }
\begin{center}
\label{tab:HIGGSCPUvsFPGATIME1282828}


\begin{tabular}{l|l|l|c|}
\cline{2-4}
\textbf{}                                         
& \textbf{Mean} & \textbf{STDV} & \textbf{Speed up} \\ \hline
\multicolumn{1}{|l|}{\textit{CPU Implementation}}  
&         5523.7 $ms$ &93.1 $ms$       &   \multirow{2}{*}{\textbf{5.9}} \\ \cline{1-3}
\multicolumn{1}{|l|}{\textit{FPGA Implementation}}
&               931.1 $ms$ & 2.3 $ms$             &                   \\ \hline
\end{tabular}
\end{center}
\end{table}

\subsubsection{4 layer network with hidden size equal to 28}
\begin{table}[H]
\def\arraystretch{1.25}%
\caption{Comparing execution time of C implementation vs FPGA implementation }
\begin{center}
\label{tab:HIGGSCPUvsFPGATIME228282828}



\begin{tabular}{l|l|l|c|}
\cline{2-4}
\textbf{}                                         
& \textbf{Mean} & \textbf{STDV} & \textbf{Speed up} \\ \hline
\multicolumn{1}{|l|}{\textit{CPU Implementation}}  
&          8437.2 $ms$ & 156.2 $ms$             &   \multirow{2}{*}{\textbf{6}} \\ \cline{1-3}
\multicolumn{1}{|l|}{\textit{FPGA Implementation}}
&             1387.2 $ms$ & 3.5$ms$   22         &                   \\ \hline
\end{tabular}

\end{center}
\end{table}

\section{Run-time Relation to Network Complexity}
\label{exp-relation}
\subsection{HIGGS}
In this section, we investigated the relation of the execution time of each training iteration with hidden size in a 3 layer neural network on a subset of HIGGS.

\begin{figure}[H]
\centering
\includegraphics[width = 0.8\hsize]{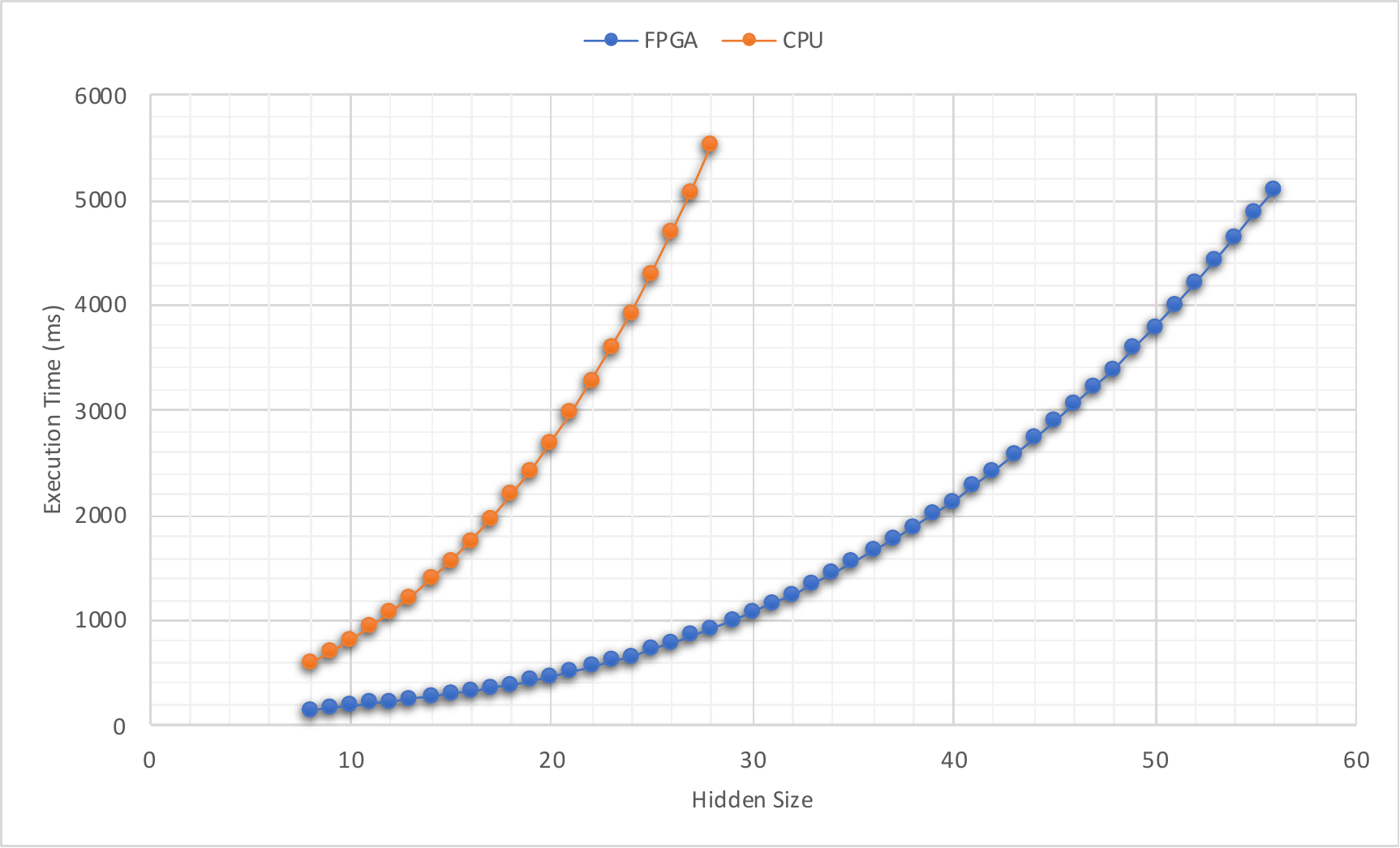}
\caption{Correlation of execution time to hidden of network}
\label{fig:HIGGSRelationFPGAHS}
\end{figure}

It is observed that while the FPGA implementation constantly performs faster, the run time of both implementations grows in a non-linear fashion when the hidden size is increased. Also, it is evident that the CPU implementation is more sensitive to hidden size changes and the gap between the execution time of implementations also grows with hidden size.
\chapter{Conclusion and Future Work}
\label{ch6}

The ADMM-LSMR method was introduced for the first time in the \cite{iso} alongside a Python implementation. It is also known that floating-point operations will cause performance issues on hardware designs. Hence we had to replace most of the arithmetic with custom implemented fixed-point arithmetic. Therefore, the feasibility of a hardware design that maintains accuracy while being deploy-able on a reasonable FPGA board was not clear at the beginning. Not only we were able to achieve such design, but also several stages of optimisation were applied to improve the initial algorithm and maximise the parallelism and hardware utilisation and achieve noticeable speed up comparing to equivalent CPU implementation. \vspace{8pt}\\
Based on the experimental results, the FPGA accelerated program was able to achieve the same accuracy as the original implementation with \textbf{\textit{less than 1\% loss using the fixed-point LSMR with nearest rounding}}. Additionally, the implementation has shown up to \textbf{\textit{6 times speed up}} depending on the network size, and architecture and it is evident that the acceleration is more effective on larger networks regarding the hidden size. 

\section{Technical Achievements }
The achievements of this project can be summarised as the following:

\begin{itemize}[label=$\sqbullet$]
     \item \textbf{C implementation}:
     ADMM-LSMR training method fully implemented in C for the first time. This was a necessity for OpenCL implementation and enabled us to perform a bottleneck analysis and identify LSMR as the target for acceleration.
     
     \item \textbf{Fixed-point implementation}: As a known technique for hardware implementations, fixed-point arithmetic with various rounding methods were implemented. Also, 16-bit and 32-bit were used with flexibility of setting the precision. These variants were employed in the LSMR module.
     \item \textbf{OpenCL implementation}: Conversion of the C implementation to an OpenCL accelerated program composed of CPU(host) program and device(FPGA) kernels. This was achieved by several structural changes both to adopt OpenCL models and Intel OpenCL SDK for FPGA  guidelines and led to successful training using the emulated FPGA.
     \item \textbf{FPGA deployment}: By getting access to Intel DevCloud environment, we were able to deploy and test the program on an actual FPGA board. This step demanded more design changes and primary optimisations as the emulation is not exactly equivalent to real hardware and hardware capacity has been added to the constrains. 
     \item \textbf{Optimisations}: There have been multiple stages of optimisation applied to the primary design. We were able to both speed up the design and reduce the utilised hardware resource in each iteration and finally fit multiple duplicates of the design on the target board to maximise utilisation and consequently the speed up.
     \end{itemize}
    \section{Results and Observations}
    The key observations of this work can be summarised as following:
     \begin{itemize}[label=$\sqbullet$]
     \item 
    
    Accuracy of ADMM-LSMR method was assessed in \cite{iso}, and it was observed that this method is able to achieve higher accuracy compared to SGD and Adam, which are two commonly used gradient-based optimisers. In this work, the accuracy of the implementation has been constantly assessed during the development both for checking the implementation correctness and more importantly, verifying the feasibility of applied techniques like variants of the fixed-point arithmetic. We were able to maintain the accuracy of the original ADMM-LSMR method with less than 1\% penalty while using nearest rounding method on HIGGS and IRIS data sets.
     \item 
     After reaching an acceptable design regarding the hardware utilisation and performance reports, the performance of the program was measured in several ways. In general, we were able to demonstrate 6 times speed up comparing to CPU implementation. We also assessed the impact of the architecture on acceleration by increasing the hidden size and observed more effectiveness on larger networks.  
     \item We observed that the nearest rounding method is more effective in the ADMM-LSMR method. This observation was unexpected as it is reported that stochastic rounding is more efficient when fixed-point arithmetic is used with gradient-based methods. The nearest rounding is simpler than stochastic rounding as it does not have the overhead of pseudo-random number generator and requires less resources. Considering this fact and our observation, the nearest rounding method was used in the final implementation.
\end{itemize}

\section{Future Work}
Some areas can  be improved and many ideas can be employed to extend this work, such as:
\begin{itemize}[label=$\sqbullet$]
    \item Design and utilisation of a 16-bit iterative least-square solver. 
    \item Using other iterative least-square solvers like LSLQ
    \item Perform more hardware optimisations and potentially increase the speed up.
    \item Using more than one devices. 
    \item Employ full or partial HDL implementation to maximise hardware utilisation and efficiency.
    \item Assess the method on other architectures of neural networks.
\end{itemize}

\bibliography{main}
\bibliographystyle{ieeetr}
\end{document}